\pdfoutput=1
\documentclass[11pt]{article}
\usepackage[colorinlistoftodos,prependcaption]{todonotes}

\usepackage[preprint]{acl}
\usepackage{times}
\usepackage{latexsym}
\usepackage{amssymb} 
\usepackage{algorithm}
\usepackage{mathtools}
\allowdisplaybreaks
\usepackage{algpseudocode} 
\usepackage{booktabs}
\usepackage{tabularx}
\usepackage{multirow}   
\usepackage{longtable}
\usepackage{amsfonts}
\newcommand{\Ind}{\mathbb{I}}
\usepackage{siunitx}
\newcommand{\vishit}{\mathrm{VIS\!-\!Hit@}k}
\newcommand{\txtmiss}{\mathrm{TXT\!-\!MissRate}}

\usepackage{booktabs}
\usepackage{caption} 
\usepackage{siunitx}
\usepackage{multirow}
\usepackage{rotating}  
\newcommand{\entails}{\vDash}        

\usepackage[T1]{fontenc}

\usepackage[utf8]{inputenc}
\usepackage{amsmath}
\usepackage{microtype}

\usepackage{inconsolata}

\usepackage{graphicx}

\title{MRAG-Suite: A Cross-Domain Diagnostic Benchmark and Tooling for Visual Retrieval-Augmented Generation}
\newcommand{\best}[1]{\textbf{#1}}           
\newcommand{\second}[1]{\underline{#1}}

\newcommand{\bestgo}{\textsuperscript{\dag}} 
\newcommand{\bestgpd}{\textsuperscript{\ddag}}
\newcommand{\robust}{\textsuperscript{\*}}   

\author{Yuelyu Ji\\
  University of Pittsburgh \\
  \texttt{yueluji@gmail.com} \\\And
  Patrick NG\\
  \texttt{patrickldld@gmail.com} \\\AND
Wuwei Lan\\
  \texttt{lwwscc@gmail.com} \\}

\begin{document}
\maketitle
\begin{abstract}
We introduce MRAG-Suite, a cross-domain evaluation suite for Visual Retrieval-Augmented Generation (RAG) that stresses real-world retrieval, multimodal grounding, and robustness. The suite consolidates eight heterogeneous sources (web photos, charts/plots, scanned documents, slides, scholarly PDFs) under unified formats and metrics. To better reflect realistic usage, we add two-step difficulty filtering and a 200-item ambiguity split, and evaluate with three retrieval modes (gold-only, gold+ distractors, distractors-only). We also present MM-RAGChecker, a claim-level diagnostic judge for multimodal grounding, reporting hallucination, faithfulness, claim recall, context precision, and cross-modality agreement/attribution. Across open-source and proprietary VLMs, we observe large drops under difficult/ambiguous queries and in the presence of distractors, as well as strong modality biases and under-utilization of retrieved evidence. MRAG-Suite and MM-RAGChecker provide actionable, fine-grained signals for improving evidence selection and justification in visual RAG systems.
\end{abstract}

\section{Introduction}

Retrieval-augmented generation (RAG) grounds large language models (LLMs) in external evidence to improve factuality and reliability. With the rise of multimodal models, \emph{visual} RAG—jointly leveraging text and images—has gained traction for complex questions involving charts, plots, figures, photos, and document pages. Yet evaluation has not kept pace: most multimodal benchmarks either assume pre-selected visuals (thus bypassing retrieval) or remain text-only, limiting our ability to measure \emph{retrieval accuracy, multimodal grounding, and robustness} in realistic settings.

Existing multimodal QA datasets such as ScienceQA~\citep{Lu2022LearnTE}, MMMU~\citep{Yue2023MMMUAM}, and MMMU-Pro~\citep{Yue2024MMMUProAM} supply the necessary visual context in the input, side-stepping retrieval. Retrieval-centric efforts like Visual Haystacks~\citep{Wu2024VisualHA} emphasize hard look-up but focus on a single domain and lack fine-grained diagnosis. Text-focused diagnostic tools (e.g., RAGChecker, RefChecker)~\citep{ru2024ragchecker, Hu2024RefCheckerRF} provide claim-level analysis, but cannot assess how claims are grounded in \emph{visual} evidence.

\begin{figure}[t]
  \centering
  \includegraphics[width=0.45\textwidth]{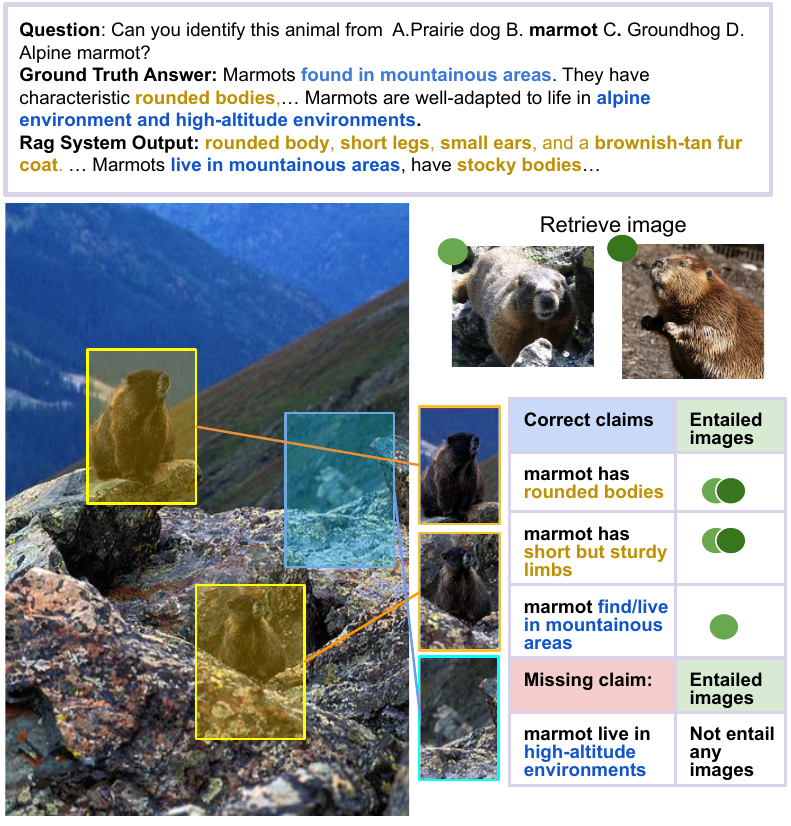}
  \caption{Illustration of our multimodal claim verification process.
  Given a long-form answer from a RAG system, we extract atomic claims and assess support from retrieved visual evidence.
  Claims are labeled \textbf{Entailment} if at least one image supports them. In this example, several traits are correctly described, but the gold claim about ``high-altitude environments'' remains unsupported.}
  \label{fig:framework}
\end{figure}

To bridge these gaps, we introduce \textbf{MRAG-Suite}, a cross-domain diagnostic evaluation suite for \emph{visual} RAG. MRAG-Suite integrates eight heterogeneous sources spanning charts/plots, natural photos, scanned documents, slides, and scholarly PDFs, and standardizes them under unified formats and metrics. To reflect real-world usage, the suite (i) applies difficulty-based filtering to de-trivialize questions, (ii) builds a 200-item ambiguity split to test underspecified queries, and (iii) evaluates three retrieval modes (gold-only, gold+ distractors, distractors-only) to probe robustness against distractors. 

We further present \textbf{MM-RAGChecker}, a claim-level multimodal judge that verifies long-form answers against retrieved \emph{text} and \emph{images}. It reports hallucination and faithfulness alongside claim recall, context precision, and cross-modality behavior (e.g., agreement and attribution), yielding actionable diagnostics for error analysis and model development.

\paragraph{Contributions.}
\begin{enumerate}
\item \textbf{A unified, multi-domain visual RAG benchmark.} We consolidate and normalize datasets across diverse domains, providing unified evaluation standards and enabling direct comparisons of model performance across various multimodal tasks.
\item \textbf{Difficulty- and ambiguity-aware evaluation.} Two-step filtering removes trivial items; an ambiguity subset stresses underspecified queries. Controlled distractor settings quantify robustness.
\item \textbf{Claim-level multimodal diagnostics.} MM-RAGChecker extends textual claim checking~\citep{ru2024ragchecker, Hu2024RefCheckerRF} to visual contexts, attributing support or errors to specific modalities and retrieved items.
\end{enumerate}

All code, prompts, and diagnostics are available at: \\
\texttt{\url{https://anonymous.4open.science/status/MRAGChecker-B33D}}

\section{Related Work}

\paragraph{Retrieval-Augmented Generation Evaluation} Traditional evaluation of RAG systems primarily relied on metrics like exact match and ROUGE scores, which are insufficiently fine-grained to diagnose specific failures. RAGChecker \citep{ru2024ragchecker} introduced claim-level diagnostics for textual RAG, evaluating retrieval accuracy, factual consistency, and generation hallucinations. However, these tools remain limited to text-only contexts, lacking the capability to assess visual evidence integration and multimodal reasoning.

\paragraph{Multimodal QA and Reasoning Benchmarks} Benchmarks such as ScienceQA \citep{Lu2022LearnTE}, MMMU \citep{Yue2023MMMUAM}, and their advanced variants MMMU-Pro \citep{Yue2024MMMUProAM} evaluate models’ abilities to reason jointly over provided image and text inputs but exclude retrieval challenges. In contrast, Visual Haystacks \citep{Wu2024VisualHA} targets multimodal retrieval tasks but is constrained to single-domain visual reasoning, omitting extensive claim-level diagnostics and failing to capture multimodal reasoning intricacies across diverse domains\cite{li2025human,li2025chatmotion}.

\paragraph{Biases in Multimodal Retrieval-Augmented Systems} Recent studies reveal significant biases in multimodal RAG systems. For example, Yao et al. \citep{Yao2025WhoII} demonstrated strong positional and modality biases, with models disproportionately relying on initial evidence items or textual over visual modalities. However, existing benchmarks have not systematically quantified these biases across multiple domains under controlled experimental conditions. MRAG-Suite specifically addresses this shortcoming by incorporating varied distractor and ambiguity scenarios to measure these biases explicitly\cite{wang2025selfdestructivelanguagemodel,liang2025graphrag,jiang2024robustkv,dan2024evaluation,lu2025predicting,li2025frequency}.

\paragraph{Domain-specific Multimodal QA} Prior work on multimodal QA has predominantly focused on specific domains or modalities, such as ChartMRAG \citep{yang2025benchmarking} and VDocRAG \citep{Tanaka2025VDocRAGRG}, demonstrating the feasibility and utility of retrieval-augmented QA systems in targeted contexts. Nonetheless, these domain-specific evaluations lack a unified framework for cross-domain comparisons and fine-grained diagnostics across modalities, a gap that MRAG-Suite explicitly aims to fill.

\section{MRAG-Suite Benchmark Construction}
\label{sec:construction}

\subsection{Task Definition}
We evaluate multimodal Retrieval-Augmented Generation (RAG) for open-ended QA. 
Given a query $q$, the system retrieves textual ($E^{\text{txt}}$) and visual ($E^{\text{img}}$) evidence and must produce (i) a concise short answer and (ii) a grounded long answer:
\(
(a^{\text{short}}, a^{\text{long}}) = f\!\big(q, E^{\text{txt}}, E^{\text{img}}\big).
\)

\subsection{Corpora Overview}
\label{sec:corpora}
MRAG-Suite merges eight heterogeneous sources spanning charts, natural photos, open-web images, scanned PDFs, and slides. 
Each source contributes distinct modalities and reasoning demands:

\begin{itemize}
  \item \textbf{WebQA}~\cite{chang2022WebQA}: open-web questions with both images (captions/descriptions) and short text snippets; often multi-hop and open-ended.
  \item \textbf{Chart-RAG}~\cite{yang2025benchmarking}(from ChartQA~\cite{masry2022chartqa}): chart understanding (e.g., ``When did X peak?''); evidence includes the chart image plus OCR/metadata; answers are numeric/categorical values.
  \item \textbf{Visual-RAG~\cite{Wu2025VisualRAGBT}}: queries that \emph{require} visual knowledge (object/attribute recognition), drawn from Visual7W/FVQA/MRAG-Bench-style scenarios.
  \item \textbf{MRAG-Bench}~\cite{hu2024mrag}: 1,353 vision-centric MCQ converted to open-ended by taking the correct option as $a^{\text{short}}$; emphasizes fine-grained and multi-view recognition.
  \item \textbf{VisRAG}~\cite{Yu2024VisRAGVR} subsets: Arxiv~\cite{Li2024MultimodalAA} (scholarly PDF pages), Plot~\cite{Methani2019PlotQARO} (technical plots), Slide~\cite{10.1609/aaai.v37i11.26598} (slide decks), and Document~\cite{Tito2022HierarchicalMT} (scanned manuals/forms). Each page is treated as a visual document with OCR text.
  
\end{itemize}

\paragraph{Dataset statistics.}
Table~\ref{tab:dataset-stats} summarizes domains, filtering counts, image resolutions, and text length statistics.

\begin{table*}[t]
  \centering
  \small
  \setlength{\tabcolsep}{4pt}
  \caption{Overview of the eight sources in MRAG-Suite, covering domains, filtering counts, image resolution, and question/answer lengths. The suite spans open-web images, charts/plots, scanned documents, slides, and scholarly PDFs to enable cross-domain visual RAG evaluation.}

  \label{tab:dataset-stats}
  \begin{tabularx}{\textwidth}{l X c X c c c c c}
    \toprule
    \textbf{Dataset}
      & \textbf{Domain}
      & \textbf{Two-way filter}
      & \textbf{Avg.\ Img (px)}
      & \textbf{Max Q}
      & \textbf{Max A}
      & \textbf{Avg Q}
      & \textbf{Avg A}
      & \textbf{Ambig.\#} \\
    \midrule
    MRAG-Bench & Natural photos (animals, objects) & 845/1353 & 803$\times$658 & 20 & 9 & 8 & 2 & 753/1353 \\
    Chart-RAG  & Business / science charts         & 2720/4738 & 466$\times$576 & 74 & 81 & 7 & 20.7 & 4093/4738 \\
    Visual-RAG & Fine-grained natural images       & 298/378    & 429$\times$430 & 39 & 34 & 19 & 5.8 & 309/378 \\
    WebQA      & Open-web images \& captions       & 19480/21382 & 792$\times$641 & 70 & 99 & 17.5 & 12.5 & 9730/21382 \\
    VisRAG-ArXiv & Scholarly PDF pages             & 709/816   & 1790$\times$1278 & 40 & 21 & 22 & 12 & 46/816 \\
    VisRAG-Plot  & Technical plots                  & 794/863   & 1098$\times$684 & 45 & 4  & 15 & 8  & 9/863 \\
    VisRAG-Slide & Slide decks (mixed media)        & 525/556   & 1026$\times$734 & 39 & 15 & 14 & 10 & 101/556 \\
    VisRAG-Doc   & Scanned manuals / forms          & 562/591   & 1845$\times$2244 & 24 & 37 & 20 & 4  & 6/591 \\
    \bottomrule
  \end{tabularx}
\end{table*}

\begin{figure*}[t]
  \centering
  \includegraphics[width=\textwidth]{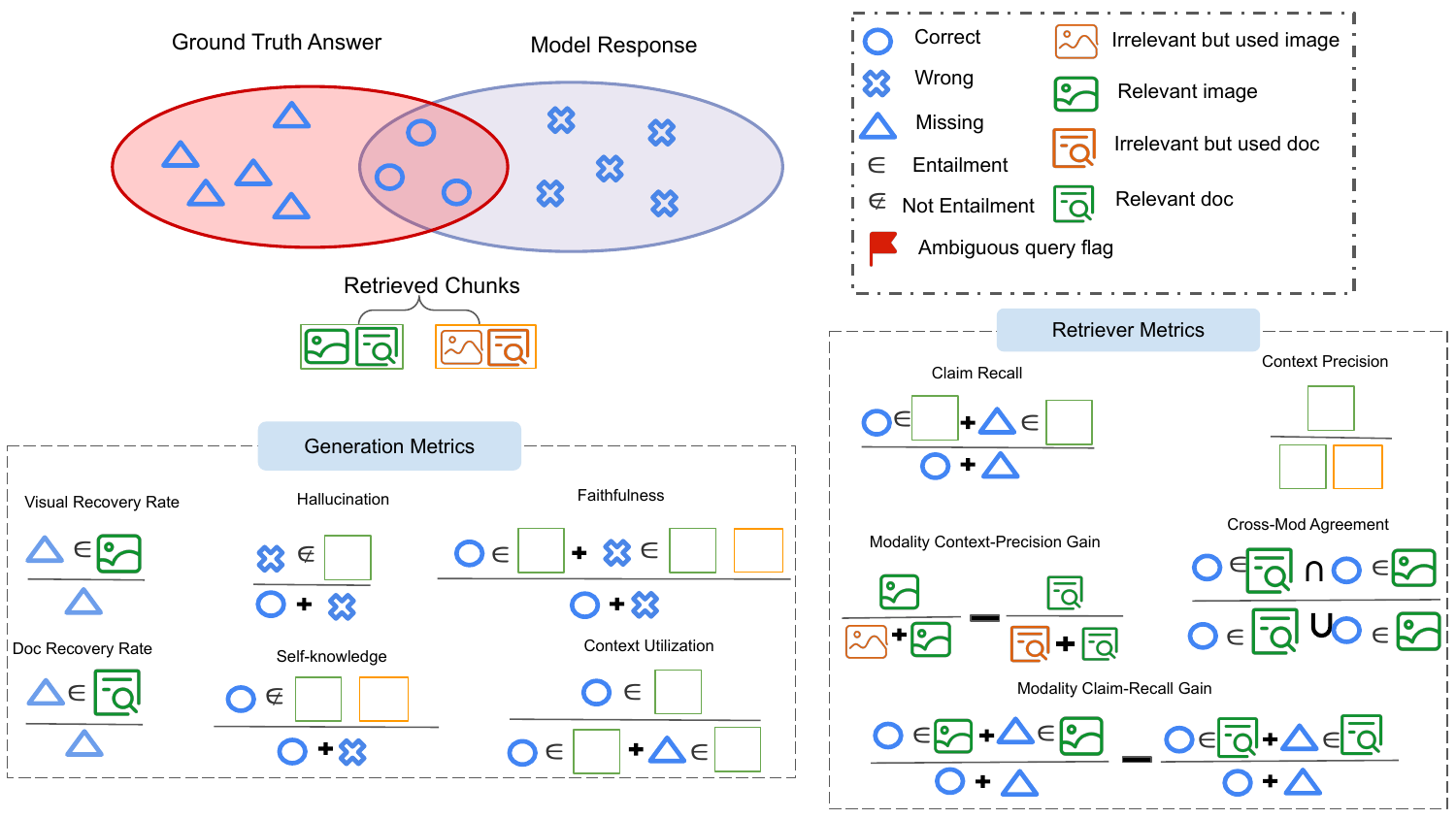}
  \caption{MM-RAGChecker overview: claim extraction, multimodal evidence matching, and verification with per-claim labels and diagnostic metrics.}

  \label{fig:metrix}
\end{figure*}
\subsection{Normalization and Indexing}
All samples are normalized to
\texttt{\{question, short\_answer, long\_answer, evidence\_imgs, evidence\_txts\}}.
Images use CLIP (ViT-L/14) features plus captions\cite{radford2021learning}; text passages use dense embeddings.
Visual items are indexed in FAISS; text is stored in a DPR/ElasticSearch store.

\subsection{Two-Step Filtering for Non-triviality}
We remove trivial and guessable items via:
\begin{enumerate}
  \item \textbf{Retrieval-Independent Filter}: questions solved by a strong closed-book model (Claude 3.5, confidence $>0.9$) or whose answers appear verbatim in the question.
  \item \textbf{Difficulty-Based Filter}: rank remaining items by multi-hop requirement, modality dependency, and baseline success rate; drop the easiest $\sim$10\% per domain.
\end{enumerate}
We release two query variants: \textit{Filtered} (minimal, disambiguated) and \textit{Full} (original, potentially noisy).
\paragraph{Ambiguity-aware evaluation.}
To isolate the impact of query ambiguity on Visual RAG, we construct an \textit{ambiguity-aware} subset (200 items) using a two-stage pipeline: an LLM-based pre-filter with short rationales, followed by double-annotator adjudication (substantial agreement, $\kappa\!=\!0.74$). We also provide evidence-grounded answer rewriting and disambiguated query rewrites for these items. Full prompts and guidelines are deferred to Appendix~\ref{app:ambiguity_full}.

\subsection{Result of Distractors}
To probe robustness, we evaluate three settings:
\begin{itemize}
  \item \textbf{gt\_only (GO)}: only gold evidence;
  \item \textbf{gt\_plus\_distractors (GPD)}: gold mixed with CLIP-similar distractors;
  \item \textbf{distractors\_only (DO)}: only distractors, testing spurious reasoning.
\end{itemize}

\subsection{Output Specification}
Systems must return both short and long answers.
We compute EM/Accuracy for $a^{\text{short}}$, and claim-level diagnostics plus ROUGE-L for $a^{\text{long}}$ (MM-RAGChecker in Sec.~\ref{sec:mmragchecker}).

Having standardized data and retrieval settings, we next introduce \textbf{MM-RAGChecker}, a claim-level diagnostic tool that evaluates how long-form answers are grounded in the retrieved multimodal evidence. It operates in three stages—claim extraction, evidence matching, and verification—and produces a set of fine-grained metrics for hallucination, faithfulness, and cross-modality behavior.
\paragraph{Evaluation axes (metric families).}
We evaluate along two complementary metric families rather than by answer length:
(i) \textbf{End-task accuracy} on the short answer ($a^{\text{short}}$) via EM/Acc; and
(ii) \textbf{Claim-level diagnostics} on the long answer ($a^{\text{long}}$) using MM-RAGChecker
(\S\ref{sec:mmragchecker}; Hallucination, Faithfulness, Claim Recall, Context Precision, Self-Knowledge, and cross-modality metrics).

\section{MM-RAGChecker: Multimodal Claim-Level Diagnosis}
\label{sec:mmragchecker}
\subsection{Overview}
\textbf{Stage 1 — Claim extraction.}
We split the long answer $a^{\text{long}}$ into minimal verifiable units
$\mathcal{C}=\{c_i\}$ using simple heuristics plus an LLM splitter for
compound clauses (e.g., numeric comparisons).

\textbf{Stage 2 — Per-evidence judging (3-way).}
Given images $\{I_k\}_{k=1}^{K_{\mathrm{img}}}$ and text passages
$\{t_k\}_{k=1}^{K_{\mathrm{txt}}}$ retrieved for the question, we directly
judge each pair $(c_i,e)$, $e\in\{I_k\}\cup\{t_k\}$, with a three-way decision
$\{\textsc{Entailment},\textsc{Neutral},\textsc{Contradiction}\}$.
No image captions are synthesized and no numeric scores are used.
We cap to top $K_{\mathrm{img}}{=}3$ and $K_{\mathrm{txt}}{=}3$ by default and
set temperature to zero for determinism.
For each claim we store two multisets of judgments
$\mathcal{J}^{\mathrm{img}}_i=\{j(c_i,I_k)\}$ and
$\mathcal{J}^{\mathrm{txt}}_i=\{j(c_i,t_k)\}$.

\textbf{Stage 3 — Aggregation to claim labels.}
Each claim receives a single final label $L_i\in
\{\textsc{Entailment},\textsc{Neutral},\textsc{Contradiction}\}$ via a simple
precedence rule:
(i) $L_i{=}\textsc{Entailment}$ if any $j\in
\mathcal{J}^{\mathrm{img}}_i\cup\mathcal{J}^{\mathrm{txt}}_i$ is
\textsc{Entailment};
(ii) else $L_i{=}\textsc{Contradiction}$ if any judgment is
\textsc{Contradiction};
(iii) else $L_i{=}\textsc{Neutral}$.
For modality analysis, define binary flags
$s^{\mathrm{img}}_i{=}1$ iff some $j\in\mathcal{J}^{\mathrm{img}}_i$ is
\textsc{Entailment}, and analogously $s^{\mathrm{txt}}_i$ for text.
We also mark a retrieved item $e$ as \emph{used} if there exists a claim
$c_i$ with $j(c_i,e){=}\textsc{Entailment}$.
All metrics below are computed on these final labels and flags.

\subsection{Metrics}

\paragraph{Core metrics.}
Let $\mathcal{C}$ be the set of extracted claims, $L_i$ the final label of
$c_i$, and $\mathcal{E}$ the set of retrieved items. Then

\begingroup\small             
\begin{align}
  \mathrm{Hallucination\ Rate}
  &= \frac{\#\{\,i : L_i=\textsc{Neutral}\,\}}{|\mathcal{C}|}, \\
  \mathrm{Faithfulness}
  &= \frac{\#\{\,i : L_i=\textsc{Entailment}\,\}}{|\mathcal{C}|}.
\end{align}
\endgroup

Let $\mathcal{G}$ be gold salient claims mined from a reference long answer.

\begingroup
\small
\setlength{\abovedisplayskip}{6pt}
\setlength{\belowdisplayskip}{6pt}
\setlength{\jot}{2pt} 
\begin{align}
  \mathrm{Claim\ Recall\ (CR)}
  &= \frac{\#\{\text{gold } g\in\mathcal{G}\ \text{that appear in } a^{\text{long}}\}}{|\mathcal{G}|}.
\end{align}
\endgroup

Let $\mathcal{E}_{\mathrm{used}}=\{e\in\mathcal{E}:\exists i,\ j(c_i,e)=\textsc{Entailment}\}$.

\begingroup\small
\begin{align}
  \mathrm{Context\ Precision\ (CP)}
  &= \tfrac{|\mathcal{E}_{\mathrm{used}}|}{|\mathcal{E}|}.
\end{align}
\endgroup

Self-knowledge measures how many entailed claims did not rely on retrieval:

\begingroup \small
\begin{align}
\resizebox{\linewidth}{!}{$
  \mathrm{Self\text{-}Knowledge}
  = \frac{\#\{\textsc{Entailment } c_i\ \text{with no } j(c_i,e)=\textsc{Entailment}\ \forall e\in\mathcal{E}\}}
           {\#\{\textsc{Entailment } c_i\}}
$}
\end{align}
\endgroup

\paragraph{Cross-modality metrics.}
We quantify cross-modality behavior with:
\emph{Claim Recall (CR)}, \emph{Claim Precision (CP)}, \emph{Visual Hit@}$k$ (\emph{VIS-Hit@}$k$),
\emph{Text Miss Rate} (\emph{TXT-MissRate}),
\emph{Cross-Modal Agreement} (\emph{CMA}),
\emph{Vision-attributed Hallucination Rate} (\emph{V-HR}),
and \emph{Document/Text-attributed Hallucination Rate} (\emph{D-HR}).
Let $\mathcal{C}$ be the set of generated \emph{atomic claims} for a sample and
$L_i \in \{\textsc{Entailment},\textsc{Contradicted},\textsc{Neutral}\}$ the
verdict from MM-RAGChecker for claim $i\in\mathcal{C}$ with respect to the provided context.
Let $\mathcal{I}$ and $\mathcal{T}$ denote the retrieved image and text sets; we write $v\models i$ if evidence $v$ entails claim $i$.

\begingroup\small
\setlength{\abovedisplayskip}{6pt}\setlength{\belowdisplayskip}{6pt}
\begin{align}
  s^{\mathrm{img}}_i &= \mathbf{1}\{\,\exists v\in\mathcal{I}: v \models i\,\}, \label{eq:simg}\\
  s^{\mathrm{txt}}_i &= \mathbf{1}\{\,\exists t\in\mathcal{T}: t \models i\,\}. \label{eq:stxt}
\end{align}
\endgroup

Claim-level recall/precision for $m\in\{\mathrm{img},\mathrm{txt}\}$:
\begingroup\small
\setlength{\abovedisplayskip}{6pt}\setlength{\belowdisplayskip}{6pt}
\begin{align}
  \mathrm{CR}_m &=
  \frac{\sum_{i\in\mathcal{C}}\mathbf{1}[L_i=\textsc{Entailment}]\,\mathbf{1}[s^m_i=1]}
       {\sum_{i\in\mathcal{C}}\mathbf{1}[L_i=\textsc{Entailment}]},
  \label{eq:crm}\\
  \mathrm{CP}_m &=
  \frac{\sum_{i\in\mathcal{C}}\mathbf{1}[s^m_i=1]}
       {\sum_{i\in\mathcal{C}}\mathbf{1}[\,s^{\mathrm{img}}_i=1 \vee s^{\mathrm{txt}}_i=1\,]}.
  \label{eq:cpm}
\end{align}
\endgroup

Cross-modality gaps:
\begingroup\small
\setlength{\abovedisplayskip}{6pt}\setlength{\belowdisplayskip}{6pt}
\begin{align}
  \Delta\mathrm{CR} &= \mathrm{CR}_{\mathrm{img}}-\mathrm{CR}_{\mathrm{txt}}, \label{eq:dcr}\\
  \Delta\mathrm{CP} &= \mathrm{CP}_{\mathrm{img}}-\mathrm{CP}_{\mathrm{txt}}. \label{eq:dcp}
\end{align}
\endgroup

Coverage (restricting to top-$k$ per modality):

\begingroup\small
\setlength{\abovedisplayskip}{4pt}\setlength{\belowdisplayskip}{6pt}
\begin{align}
  \vishit &= \frac{1}{k}\sum_{i=1}^{k} \Ind\!\big[\,I_i \entails \exists c\,\big], \label{eq:vishit}\\
  \txtmiss &= 1 - \frac{1}{k}\sum_{i=1}^{k} \Ind\!\big[\,t_i \entails \exists c\,\big]. \label{eq:txtmiss}
\end{align}
\endgroup
Cross-modal agreement:
\begingroup\small
\setlength{\abovedisplayskip}{6pt}\setlength{\belowdisplayskip}{6pt}
\begin{equation}
  \mathrm{CMA} =
  \frac{\sum_{i\in\mathcal{C}}\mathbf{1}[\,s^{\mathrm{img}}_i=1 \wedge s^{\mathrm{txt}}_i=1\,]}
       {\sum_{i\in\mathcal{C}}\mathbf{1}[\,s^{\mathrm{img}}_i=1 \vee s^{\mathrm{txt}}_i=1\,]}.
  \label{eq:cma}
\end{equation}
\endgroup

Hallucination attribution:
\begingroup\small
\setlength{\abovedisplayskip}{6pt}\setlength{\belowdisplayskip}{6pt}
\begin{align}
  \mathrm{V\!-\!HR} &=
  \frac{\#\{\,i:\ s^{\mathrm{img}}_i=1 \ \wedge\  L_i\neq \textsc{Entailment}\,\}}
       {\#\{\,i:\ L_i\neq \textsc{Entailment}\,\}},
  \label{eq:vhr}\\
  \mathrm{D\!-\!HR} &=
  \frac{\#\{\,i:\ s^{\mathrm{txt}}_i=1 \ \wedge\  L_i\neq \textsc{Entailment}\,\}}
       {\#\{\,i:\ L_i\neq \textsc{Entailment}\,\}}.
  \label{eq:dhr}
\end{align}
\endgroup

Unless noted, all values are percentage points on the \textbf{Filtered} split,
and we use $k=5$ per modality at test time.

\begin{table*}[ht]
 \centering
 \small
 \begin{tabular}{l cc cc cc cc cc cc}
   \toprule
   \multirow{2}{*}{\bf Dataset}
     & \multicolumn{2}{c}{\bf Qwen2.5-VL-72B}
     & \multicolumn{2}{c}{\bf LLaVa-v1.6-34B}
     & \multicolumn{2}{c}{\bf Claude 3.5}
     & \multicolumn{2}{c}{\bf Phi-4}
     & \multicolumn{2}{c}{\bf Pixtral-12B-2409}
     & \multicolumn{2}{c}{\bf InternVL3-8B} \\
   \cmidrule(lr){2-3}\cmidrule(lr){4-5}
   \cmidrule(lr){6-7}\cmidrule(lr){8-9}
   \cmidrule(lr){10-11}\cmidrule(lr){12-13}
     & {Filt} & {Full}
     & {Filt} & {Full}
     & {Filt} & {Full}
     & {Filt} & {Full}
     & {Filt} & {Full}
     & {Filt} & {Full} \\
\midrule
Chart-MRAG
  & \best{22.4} & 19.1
  & 15.3 & \second{25.7}
  & 10.8 & 11.9
  & 11.2 & 13.5
  & 14.7 & \best{29.4}
  & \second{19.2} & 27.6 \\
\cmidrule(lr){1-13}
MRAG-Bench
  & \second{32.6} & \second{33.8}
  & 19.5 & 24.6
  & 28.3 & 26.1
  & 32.4 & 30.7
  & \best{38.8} & \best{37.5}
  & 21.9 & 21.4 \\
\cmidrule(lr){1-13}
VisRAG-ArXiv
  & \second{55.3} & \second{48.7}
  & 34.6 & 34.5
  & 35.5 & 35.1
  & 38.6 & 36.9
  & 43.1 & 43.7
  & \best{63.9} & \best{55.4} \\
\cmidrule(lr){1-13}
VisRAG-Doc
  & 31.1 & \best{35.4}
  & 20.2 & 21.9
  & 28.5 & 29.2
  & 27.6 & 28.2
  & 26.9 & 26.6
  & \best{39.7} & \second{34.6} \\
\cmidrule(lr){1-13}
VisRAG-Plot
  & 36.7 & 45.8
  & 12.3 & 13.6
  & 35.9 & 36.8
  & 6.4 & 9.2
  & \best{51.4} & \best{56.5}
  & \second{33.5} & \second{36.1} \\
\cmidrule(lr){1-13}
VisRAG-Slide
  & 18.5 & \best{25.3}
  & 8.7 & 10.4
  & 17.6 & 22.1
  & \best{24.3} & 22.6
  & 18.2 & 15.5
  & \second{20.8} & \second{22.9} \\
\cmidrule(lr){1-13}
Visual-RAG
  & \best{13.8} & 13.1
  & 9.6 & 11.5
  & 11.7 & 10.6
  & 8.5 & 9.2
  & \second{12.1} & \best{13.3}
  & 11.2 & \second{12.6} \\
\cmidrule(lr){1-13}
WebQA
  & 16.2 & 14.9
  & 12.7 & 13.4
  & 14.8 & 17.3
  & 20.6 & 22.1
  & \second{26.4} & \second{27.9}
  & 17.8 & \best{28.7} \\
\bottomrule
\end{tabular}
\caption{EM/Accuracy performance of each model on the \textbf{Filtered} (retained valid samples) and \textbf{Full} (all samples) splits. \best{Best} and \second{Second-best} are highlighted per row and split.}
\label{tab:acc_splits}
\end{table*}

\begin{table*}[ht]
  \centering
  \scriptsize
  \setlength{\tabcolsep}{8pt}
  \begin{tabular}{l l c c c c c c c c}
  \toprule
    \multirow{2}{*}{Dataset} & \multirow{2}{*}{Model} &
    Recall & Precision & F1 & Halluc. & Faith. & Self-know. & Claim R. & Ctx.Prec.\\
    \cmidrule(lr){3-10}
     & & \scriptsize(img/txt) & \scriptsize(img/txt) & \scriptsize(img/txt) & \scriptsize(img/txt) & \scriptsize(img/txt) & \scriptsize(img/txt) & \scriptsize(img/txt) & \scriptsize(img/txt)\\
    \midrule
    \multirow{3}{*}{Chart-MRAG}
      & Qwen2.5-VL-72B   & 12.1/12.9 & 7.9/8.1 & \best{6.3}/\second{7.2} & \second{57.3}/\best{58.9} & 54.9/7.8  & 8.3/7.7  & 8.9/14.7 & 85.2/\best{31.8} \\
      & LLaVa-v1.6-34B   & 7.6/10.1  & 0.6/18.9 & 1.1/\best{13.2}       & 57.4/66.8               & 50.5/8.8  & 3.6/10.6 & 2.0/12.8 & \second{86.1}/\second{22.3} \\
      & Claude 3.5       & 13.3/13.2 & 0.8/2.1  & \second{1.4}/3.4       & \best{24.9}/87.3        & 65.1/10.6 & 0.0/2.1  & 75.0/10.0    & \best{90.0}/15.0 \\
    \addlinespace[2pt]
    \multirow{3}{*}{MRAG-Bench}
      & Qwen2.5-VL-72B   & 16.3   & 1.2   & \second{2.2} & \best{45.2} & 62.6   & 12.4  & 11.3  & 72.1   \\
      & LLaVa-v1.6-34B   & 10.9   & 1.0   & 1.8          & 57.8        & 54.8   & 3.6   & 8.5   & \best{87.4}   \\
      & Claude 3.5       & 53.3   & 4.1   & \best{7.6}   & \second{48.2} & 66.3   & 5.7   & 11.0  & \second{79.6}   \\
    \addlinespace[2pt]
    \multirow{3}{*}{VisRAG-ArXiv}
      & Qwen2.5-VL-72B   & 23.7  & 1.9   & \best{3.5}   & \best{30.3} & 67.2   & 17.2  & 23.2  & 61.0   \\
      & LLaVa-v1.6-34B   & 19.7  & 1.7   & \second{3.1} & \second{37.0} & 65.5   & 12.3  & 19.4  & \second{67.4}   \\
      & Claude 3.5       & 17.0  & 1.6   & 2.9          & 41.4        & 64.4   & 9.1   & 16.9  & \best{71.6}   \\
    \addlinespace[2pt]
    \multirow{3}{*}{VisRAG-Doc}
      & Qwen2.5-VL-72B   & 16.8  & 1.5   & \best{2.8}   & \best{45.0} & 57.5   & 8.0   & 11.1  & 74.3   \\
      & LLaVa-v1.6-34B   & 11.0  & 0.9   & 1.6          & 54.3        & 56.3   & 4.2   & 5.6   & \best{82.3}   \\
      & Claude 3.5       & 12.1  & 1.2   & \second{2.2} & \second{51.6} & 57.7   & 4.7   & 7.3   & \second{80.4}   \\
    \addlinespace[2pt]
    \multirow{3}{*}{VisRAG-Plot}
      & Qwen2.5-VL-72B   & 17.7  & 1.9   & \best{3.4}   & \best{42.1} & 65.5   & 9.6   & 16.4  & 75.1   \\
      & LLaVa-v1.6-34B   & 9.5   & 0.8   & 1.5          & 60.9        & 50.1   & 3.2   & 1.9   & \best{88.6}   \\
      & Claude 3.5       & 17.1  & 1.5   & \second{2.8} & \second{46.7} & 62.8   & 8.3   & 13.5  & \second{75.9}   \\
    \addlinespace[2pt]
    \multirow{3}{*}{VisRAG-Slide}
      & Qwen2.5-VL-72B   & 8.7   & 0.6   & \best{1.1}   & \best{52.7} & 47.7   & 4.4   & 3.2   & 85.9   \\
      & LLaVa-v1.6-34B   & 6.3   & 0.5   & \second{0.9} & 65.8        & 46.3   & 1.2   & 1.3   & \best{92.2}   \\
      & Claude 3.5       & 8.2   & 0.5   & \second{0.9} & \second{55.8} & 51.3   & 2.1   & 1.0   & \second{89.6}   \\
    \addlinespace[2pt]
    \multirow{3}{*}{Visual-RAG}
      & Qwen2.5-VL-72B   & 6.5   & 0.5   & \second{1.0} & \best{74.0} & 42.3   & 2.7   & 5.2   & 11.0   \\
      & LLaVa-v1.6-34B   & 5.3   & 0.4   & 0.8          & \second{76.8} & 44.1   & 3.2   & 4.5   & \second{12.0}   \\
      & Claude 3.5       & 6.0   & 0.6   & \best{1.1}   & 85.1        & 39.5   & 1.8   & 6.1   & \best{14.5}   \\
    \addlinespace[2pt]
    \multirow{3}{*}{WebQA}
      & Qwen2.5-VL-72B   & 8.8/8.1  & 0.8/16.1 & \second{1.5}/\best{10.8} & 62.8/\best{70.4} & 52.5/7.9  & 6.5/3.6  & 5.9/13.5 & \second{89.7}/\best{30.4} \\
      & LLaVa-v1.6-34B   & 6.5/7.1  & 0.7/15.8 & 1.3/\second{9.8}         & 64.5/70.8        & 50.2/6.2  & 2.5/4.1  & 1.1/12.2 & \best{91.4}/\second{28.6} \\
      & Claude 3.5       & 9.0/7.9  & 1.1/1.8  & \best{1.7}/3.7           & \best{26.6}/98.8 & 62.7/10.7 & 0.3/0.6  & 72.0/8.8 & 87.4/13.6 \\
    \bottomrule
  \end{tabular}
  
  \caption{Diagnostic metrics on the \textbf{Filtered} split (image/text pairs where applicable).\best{Best} and \second{Second-best} are highlighted per row and split.All metrics are reported as percentage points except F1 (points on [0,100]).(Self-knowledge $\rightarrow$ Self-know.; Claim Recall $\rightarrow$ Claim R.; Context Precision $\rightarrow$ Ctx.Prec.)}

  \label{tab:diag_filtered}
\end{table*}

\begin{table*}[t]
  \centering
  \tiny
  \renewcommand{\arraystretch}{0.65}
  \setlength{\tabcolsep}{1pt}
  \resizebox{\textwidth}{!}{%
  \begin{tabular}{l l c c c c c c c }
   \toprule
    Dataset & Model &
    $\Delta$CR & $\Delta$CP &
    VIS-Hit@k & TXT-MissRate &
    CMA & V-HR & D-HR \\
    \midrule
    Chart-MRAG & Phi-4 (GPD) & 54.54 & 59.50 & \best{81.38} &83.00 & \best{13.02} & \best{66.87} & \best{45.2} \\
    \midrule
    WebQA &Phi-4 (GPD) & 6.59 & 1.50 & \best{14.79} & \second{87.00} & \best{10.86} & \best{12.41} & \best{0.33} \\
    WebQA & Claude 3.5 (GPD) & 3.09 & -0.35 & \second{13.00} & \best{83.33} & \second{6.52} & \second{32.44} & \second{3.70} \\
    \bottomrule
  \end{tabular}}
    \caption{Cross-modality diagnostics: differences in claim recall/precision between images and text ($\Delta$CR/CP), visual hit rate (VIS-Hit@$k$), text miss rate, cross-modal agreement (CMA), and hallucination attribution (V-HR/D-HR). Higher CMA indicates stronger agreement between modalities.All values are percentage points; higher CMA indicates stronger agreement.
    }
  \label{tab:cross_modality}
\end{table*}

\begin{table*}[ht]
  \centering
  \scriptsize
  \resizebox{\textwidth}{!}{%
\begin{tabular}{l ccc ccc ccc ccc ccc ccc}
\toprule
\multirow{2}{*}{\bf Dataset}
  & \multicolumn{3}{c}{\bf Qwen2.5-VL-72B}
  & \multicolumn{3}{c}{\bf LLaVa-v1.6-34B}
  & \multicolumn{3}{c}{\bf Claude 3.5}
  & \multicolumn{3}{c}{\bf Phi-4}
  & \multicolumn{3}{c}{\bf InternVL3-8B}
  & \multicolumn{3}{c}{\bf Pixtral-12B-2409} \\
\cmidrule(lr){2-4}\cmidrule(lr){5-7}\cmidrule(lr){8-10}%
\cmidrule(lr){11-13}\cmidrule(lr){14-16}\cmidrule(lr){17-19}
& DO & GPD & GO   & DO  & GPD & GO   & DO  & GPD & GO   & DO  & GPD & GO   & DO  & GPD & GO   & DO  & GPD & GO \\
\midrule
Chart-MRAG
  & 22.7 & \best{25.4}\bestgpd\robust & \best{26.8}\bestgo
  & 10.3 & 18.1 & 13.9
  & 8.2  & 15.5 & 9.7
  & 15.9 & 19.8 & 17.4
  & 23.3 & 19.5 & 21.7
  & 14.8 & 14.1 & 16.6 \\
MRAG-Bench
  & 30.9 & 29.6\robust & \best{29.8}\bestgo
  & 17.8 & 16.4 & 20.7
  & 26.9 & 24.1 & 29.7
  & 2.1  & 2.4  & 27.9
  & 26.2 & 21.3 & 28.5
  & \best{33.6} & \best{29.8}\bestgpd & 28.2 \\
VisRAG-ArXiv
  & 4.4  & 4.1\robust & 4.3
  & 4.3  & \best{11.7}\bestgpd & \best{8.5}\bestgo
  & \best{7.3} & 4.1 & 5.0
  & 1.7  & 4.4 & 5.5
  & 6.8  & 5.1\robust & 5.2
  & 3.2  & 1.4 & 1.8 \\
VisRAG-Doc
  & 1.2  & 29.1 & 31.3
  & 12.9 & 20.5\robust & 20.2
  & 2.4  & 26.6 & 25.9
  & 2.3  & 21.6 & 28.3
  & 4.7  & \best{38.4}\bestgpd\robust & \best{38.1}\bestgo
  & 0.6  & 26.5 & 27.1 \\
VisRAG-Plot
  & 5.9  & 33.8 & 41.4
  & 8.4  & 12.9 & 12.6
  & 12.7 & 25.7 & 29.3
  & 6.2  & 6.7  & 32.1
  & 4.9  & 28.3\robust & 28.4
  & 0.8  & \best{51.2}\bestgpd & \best{52.0}\bestgo \\
VisRAG-Slide
  & 1.6  & 18.7 & 18.1
  & 1.9  & 8.4  & 8.2
  & 4.3  & \best{18.9}\bestgpd\robust & \best{18.8}\bestgo
  & 1.2  & 24.6 & 15.3
  & 0.9  & 17.5 & 15.7
  & 0.6  & 15.3 & 15.6 \\
Visual-RAG
  & 14.8 & 26.9 & \best{27.3}\bestgo
  & 13.6 & 18.7 & 18.1
  & 15.5 & 25.4 & 26.5
  & 15.1 & 21.4 & 20.2
  & 17.6 & \best{23.1}\bestgpd\robust & 23.2
  & 16.5 & 23.6 & 18.9 \\
WebQA
  & 17.2 & \best{15.8}\robust & 15.4
  & 12.3 & 12.6 & 13.4
  & 9.9  & 14.3 & \best{16.4}\bestgo
  & 17.5 & 7.3  & 12.8
  & 20.1 & 20.8 & 17.4
  & \best{25.6} & 25.3 & 0.9 \\
\bottomrule
\end{tabular}
}
\caption{Scores (EM preferred; Accuracy used when EM is unavailable) under the three retrieval modes: DO = \textit{distractors\_only}, GPD = \textit{gt\_plus\_distractors}, GO = \textit{gt\_only}. \dag\ marks the best in GO; \ddag\ marks the best in GPD; * marks the smallest \(|\mathrm{GO}-\mathrm{GPD}|\) (most robust to distractors).}
\label{tab:retrieval_modes}
\end{table*}

\subsection{Retrieval Modes and Distractors}
We compare \texttt{distractors\_only} (DO), \texttt{gt\_plus\_distractors} (GPD), and \texttt{gt\_only} (GO). Table~\ref{tab:retrieval_modes} shows that DO degrades performance substantially; GPD lies between DO and GO. The GPD–GO gap quantifies susceptibility to distractors.

\section{Experiments and Results}
\label{sec:exp}

We evaluate MRAG-Suite along three axes: (1) end-task accuracy on short answers, (2) factuality and grounding quality of long answers via MM-RAGChecker, and (3) robustness to query difficulty, ambiguity, and distractors.

\subsection{Experimental Setup}
\paragraph{Models.}
Our main baseline is \textbf{LLaVa-v1.6-34B} (vision-language model fine-tuned from LLaMA-2). Retrieval uses a CLIP-based dual encoder for images and a DPR retriever for text. At test time we fetch up to 5 passages and 5 images per query (when available) and serialize them (image features/captions + text) before generation. The model is prompted to output a short answer followed by a long answer.  
For comparison we also run \textbf{Claude 3.5} (API) on a subset to estimate the upper bound of current proprietary VLMs.

\paragraph{Metrics.}
For short answers we report EM/Accuracy. For long answers we compute ROUGE-L against our reference explanations and all claim-level diagnostics using MM-RAGChecker (Hallucination Rate, Faithfulness, Claim Recall, Context Precision, Self-Knowledge, Modality Bias). We further report the fraction of retrieved items actually used (\emph{evidence utilization}). Retrieval robustness is probed under three modes: \texttt{gt\_only} (GO), \texttt{gt\_plus\_distractors} (GPD), and \texttt{distractors\_only} (DO).
Unless noted, we report macro averages on the \textbf{Filtered} split with decoding temperature $=0$ and identical retriever outputs across models.




\subsection{End-task Accuracy (short-answer EM/Acc)}

Table~\ref{tab:acc_splits} reports EM/Accuracy across datasets and models.
Overall accuracies are modest and vary widely by domain.
On the \textbf{Filtered} split, the highest per-dataset scores appear on \emph{VisRAG-ArXiv} (e.g., up to \textbf{63.9} with InternVL3-8B) and \emph{VisRAG-Plot} (e.g., \textbf{51.4} with Pixtral-12B-2409), whereas \emph{Visual-RAG} and \emph{VisRAG-Slide} are the most challenging (typically in the \textbf{8--24} range across models).
\emph{WebQA} sits in the middle (best around \textbf{26.4} on the Filtered split), while \emph{Chart-MRAG} remains difficult (best Filtered around \textbf{22.4}).

Comparing \textbf{Filtered} vs.\ \textbf{Full}, the Full split is generally easier (e.g., \emph{Chart-MRAG} improves substantially for several models, such as Pixtral-12B-2409 from \textbf{14.7} to \textbf{29.4}), though the effect is not uniform across all datasets and models (e.g., \emph{VisRAG-ArXiv} sometimes drops for stronger models when moving to Full).
These trends confirm that our filtering increases non-triviality and retrieval dependence without relying on a single macro-average.

\subsection{Claim-level Diagnostics (MM-RAGChecker)}

Table~\ref{tab:diag_filtered} summarizes claim-level diagnostics on the \textbf{Filtered} split.
We observe substantial variation across domains and models rather than a single global rate.
\textbf{Hallucination} spans a wide range: it is comparatively low on document-like or chart-heavy sets (e.g., \emph{VisRAG-ArXiv} around \textbf{30--41} for major models; \emph{Chart-MRAG} can be as low as \textbf{24.9} on images for Claude~3.5) but very high on fine-grained visual recognition (\emph{Visual-RAG} often \textbf{74--85}).
Correspondingly, \textbf{Faithfulness} is highest on structured pages/plots and lowest on open-world visual recognition.

Evidence utilization and cross-modality signals further explain these gaps.
On \emph{WebQA}, \textbf{Context Precision} for images is high (e.g., \textbf{89--91}), while text exhibits much lower precision (often \textbf{28--31}), indicating that retrieved passages are frequently unused or unhelpful.
Across datasets we also see \textbf{modality skew}: models tend to commit to either text or images instead of reconciling both, which aligns with the elevated hallucination on visually challenging sets and the reduced per-claim F1 on those splits.

Overall, MM-RAGChecker reveals that (i) hallucinations remain frequent on visually demanding queries even when retrieval succeeds, (ii) many retrieved items are not credited as supporting any claim (low context precision, especially on text), and (iii) balancing visual and textual cues is a key bottleneck for long-form grounding.

\subsection{Effect of Retrieval Modes}
Table~\ref{tab:retrieval_modes} shows EM/Accuracy under \texttt{DO}, \texttt{GPD}, and \texttt{GO}.  
\texttt{DO} severely degrades performance (as expected), while \texttt{GPD} lies between \texttt{DO} and \texttt{GO}. The gap between \texttt{GPD} and \texttt{GO} quantifies the model's susceptibility to distractors; Chart-RAG and VisRAG-Doc exhibit the largest drops, revealing brittle selection of evidence.
The absolute \texttt{GPD}\,$\rightarrow$\,\texttt{GO} gain serves as a distractor-susceptibility index; Chart-RAG and VisRAG-Doc exhibit the largest gains (cf.\ Table~\ref{tab:retrieval_modes}).

\subsection{Prompt Sensitivity Study}
\label{sec:prompt_sensitivity}
\begin{figure}[t]
  \centering
  \includegraphics[width=0.45\textwidth]{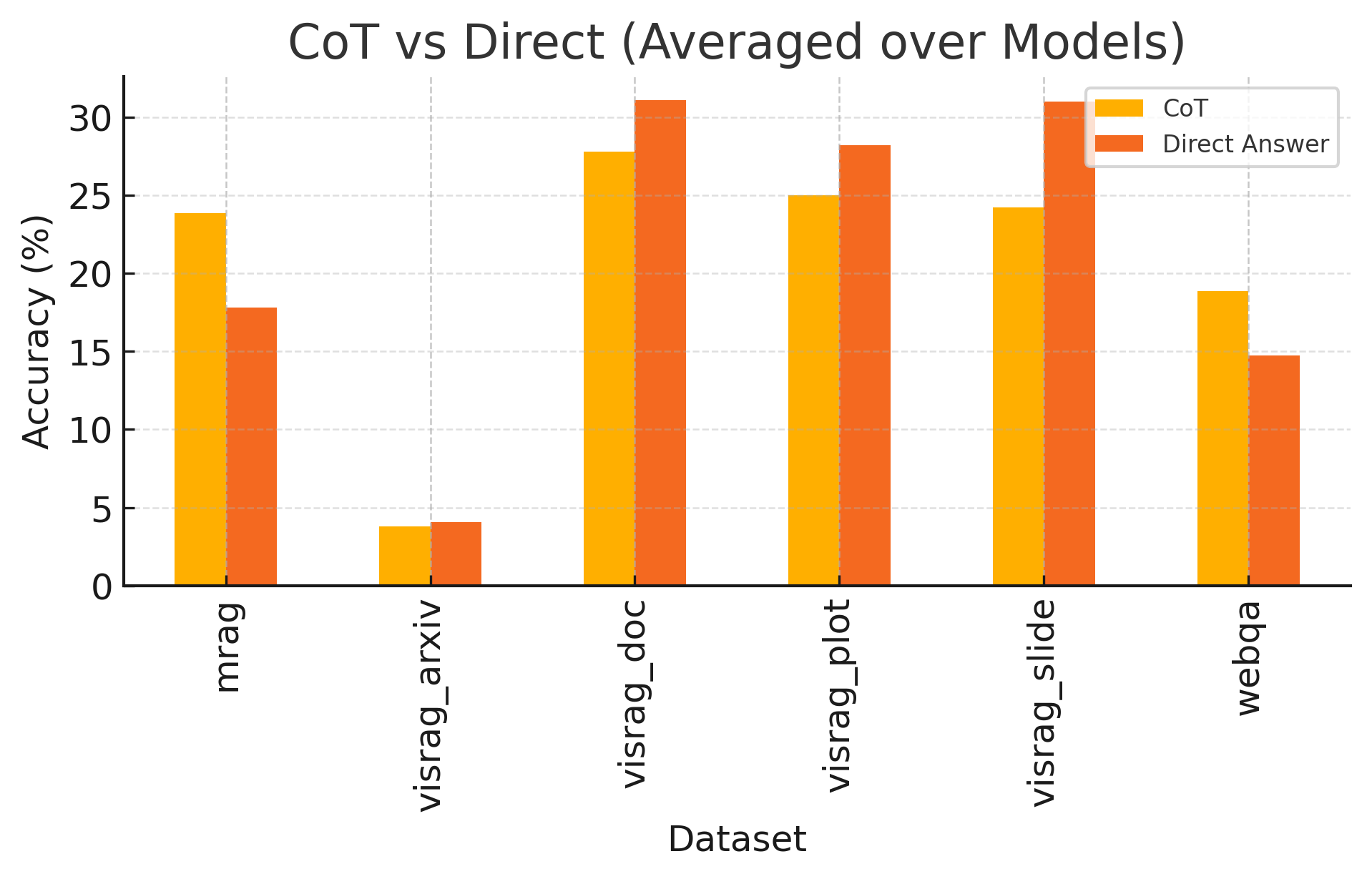}
  \caption{%
}
  \label{fig:cot_vs_direct}
\end{figure}

We sweep 12 prompt designs (init style, few-shot exemplar count, reasoning pattern). Table~\ref{tab:best_prompt} lists the best-performing prompt per dataset; Figure~\ref{fig:cot_vs_direct} visualizes the direct vs.\ retrieve-then-reason trend.
Results in Table~\ref{tab:best_prompt} are averaged over six models on the \textbf{Filtered} split with fixed seeds and decoding settings. Table~\ref{tab:best_prompt} reports, for each dataset, the prompt that achieves the highest mean Accuracy/EM across six models (filtered split). 
We observe clear patterns:
 \textbf{Reasoning matters:} \texttt{retrieve\_then\_reason} dominates on text-heavy retrieval tasks (MRAG, WebQA), whereas \texttt{direct} is consistently better on VisRAG-* visual QA splits where context is already focused.
\textbf{Few-shot count is task-dependent:} Using three exemplars (\texttt{ex3}) suffices---and sometimes outperforms \texttt{ex6}---for purely visual/numeric QA (plots/slides). In contrast, \texttt{ex6} helps more on document-style and open-web questions that benefit from pattern coverage.
 \textbf{Init style is secondary:} Switching between \texttt{plain} and \texttt{expert} seldom changes the rank unless combined with the right reasoning pattern and example count.

\begin{table}[t]
\centering
\small
\setlength{\tabcolsep}{3pt}
\begin{tabular}{l l r}
\toprule
\textbf{Dataset} & \textbf{Best Prompt} & \textbf{Mean Acc. (\%)} \\
\midrule
MRAG-Bench          & expert3 retrieve \& reason & 31.5 \\
VisRAG-ArXiv & expert6 direct                  & 5.0  \\
VisRAG-Doc   & plain6 direct                   & 34.3 \\
VisRAG-Plot  & expert3 direct                  & 36.5 \\
VisRAG-Slide & plain3 direct                   & 35.2 \\
WebQA         & expert6 retrieve\&reason  & 23.0 \\
\bottomrule
\end{tabular}

\caption{Best-performing prompt (by mean EM/Acc across models) per dataset on the filtered split. The patterns highlight when “retrieve-then-reason” helps (text-heavy tasks) versus when direct prompting is sufficient (focused visual tasks).}

\label{tab:best_prompt}
\end{table}

\subsection{Impact of Distractors on Diagnostic Metrics}
\label{sec:distractor_metrics}
Distractors hurt not only end-task EM/Acc (Table~\ref{tab:retrieval_modes}) but also grounding quality (Table~\ref{tab:diag_filtered}).
From \texttt{GO} $\rightarrow$ \texttt{GPD}/\texttt{DO}, we observe: 
(i) \textbf{higher hallucination} (unsupported clauses rise), 
(ii) \textbf{lower context precision} (more retrieved items unused or misused), and 
(iii) \textbf{stronger modality skew} (larger $|\Delta$CR$|$/|\,$\Delta$CP$|$ and lower CMA), indicating a tendency to commit to one modality under conflicting evidence.
Overall, distractors degrade both \emph{what} is answered (lower EM/Acc) and \emph{how} it is justified (higher hallucination, lower context precision, and weaker cross-modal agreement).


\subsection{Effect of Filtering}

Filtering not only reduces accuracy but also changes answer style. Filtering increases the average long-answer length from \textbf{45} to \textbf{60} tokens, yet factuality does not improve: MM-RAGChecker flags a larger fraction of clauses as unsupported on the Filtered split (cf.\ Table~\ref{tab:diag_filtered}). This suggests that harder questions trigger verbosity rather than better grounding.



\section{Conclusion}
We presented MRAG-Suite, a comprehensive suite for evaluating and diagnosing visual retrieval-augmented generation in multimodal question answering. MRAG-Suite introduces a unified benchmark that stresses truly multimodal, knowledge-intensive QA, combining diverse sources like web images, textual articles, and informational charts. To focus evaluation on the intended challenges, we devised a two-step filtering method to remove trivial cases and emphasize multi-hop, vision+language reasoning. We also contributed an automated way to generate long-form answers with evidence, enabling novel evaluation of explanations in addition to factual answers. Crucially, we developed MM-RAGChecker, a claim-level multimodal verifier, to assess the factual consistency of generated answers.

\section*{Limitations}
This work targets visual RAG for images, charts, slides, scanned pages, and scholarly PDFs; results may not transfer to videos, embodied settings, or low-resource languages. Hybrid retrieval depends on CLIP-like encoders and OCR text, so known dataset/model biases and OCR noise can affect measured recall/precision and inflate hallucination labels. Our diagnostics rely on LLM/VLM components for claim extraction and entailment; despite prompt controls and spot-checks, such judges can diverge from expert assessments in domain-specific cases. Scores are also sensitive to prompt format, retrieval $k$, and distractor sampling; we report settings but do not exhaust all combinations. Some baselines use third-party APIs, so provider updates and rate limits may introduce temporal variance despite fixed prompts and released scripts.

\bibliography{custom}
%
\appendix

\section{Ambiguity Process }

\subsection{Ambiguity Definition and Annotation}
\label{sec:ambiguity}

We curate a 200-question subset to study ambiguity in multimodal RAG.

\paragraph{Definition.}
A question is \textbf{AMBIGUOUS} if a reasonable reader cannot determine a \emph{single} correct answer without additional constraints (time, location, unit, object, visibility in the image, etc.); otherwise it is \textbf{CLEAR}. For image-conditioned queries, we additionally mark as ambiguous when key visual evidence is not visible or is blurred.

\paragraph{Detection Pipeline.}
We adopt a two-stage procedure:
\begin{enumerate}
  \item \textbf{LLM-based pre-filter.} We prompt an assessor model to output \texttt{CLEAR} or \texttt{AMBIGUOUS} together with a short rationale (30 words). We use separate prompts for text-only and image-conditioned questions (see Appendix~\ref{app:ambiguity_prompts}). 
  \item \textbf{Human adjudication.} Two annotators independently review the flagged items and resolve disagreements. We achieve substantial agreement (Cohen's $\kappa=0.74$).
\end{enumerate}
\paragraph{Rewriting and Answer Normalization.}
For each ambiguous item we:
\begin{itemize}
  \item \emph{Evidence-grounded answer rewrite}: given retrieved evidence and the original answer, an LLM produces a concise answer or returns ``Evidence inconclusive.'' (Appendix~\ref{fig:rewrite_answer}).
  \item \emph{Query disambiguation}: another prompt rewrites the question into a precise version that can be answered solely from the evidence (Appendix~\ref{fig:rewrite_query}).
\end{itemize}

The final ambiguity split (200 queries) and a matched clear set (200 queries) are used in Appendix.~\ref{sec:ambig_results} to isolate the impact of ambiguity on accuracy, claim recall, and hallucination rate.
\label{app:ambiguity_full}

\subsection{Error Patterns under Ambiguity}
\label{sec:ambig_eval}
On the 200-query ambiguity subset, two dominant patterns emerge:
\begin{enumerate}
  \item \textbf{Single-interpretation bias}: the model commits to one reasonable sense, lowering Claim Recall and short-answer accuracy.
  \item \textbf{Interpretation conflation}: retrieval returns evidence for multiple senses; the generator merges them, producing composite neutral claims and higher hallucination rates.
\end{enumerate}
These results motivate an \emph{ambiguity detection \& clarification} step before final answer generation.

\subsection{Ambiguity Detection (Image-conditioned) }
\label{app:ambiguity_prompts}
\begin{center}
  \centering
  \includegraphics[width=0.45\textwidth]{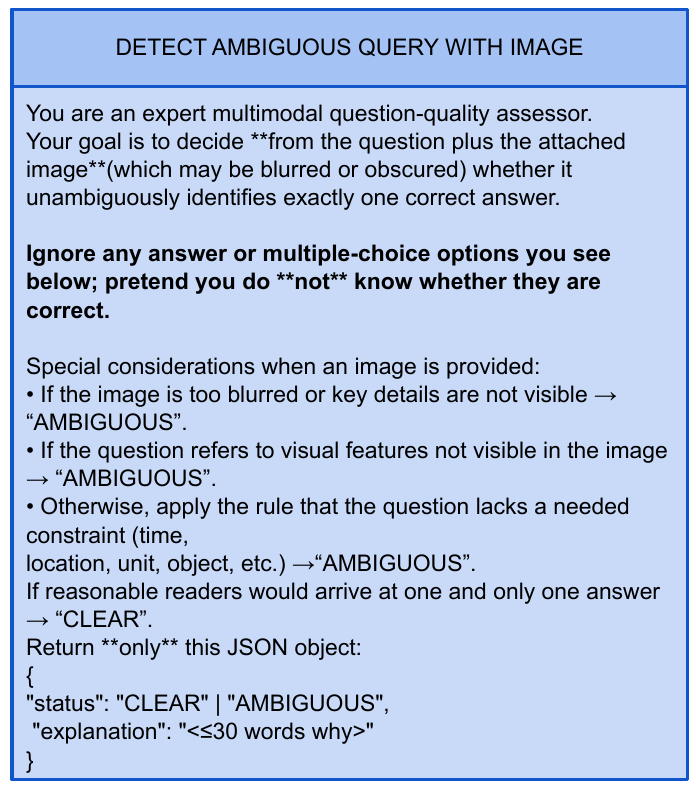}
  \label{fig:ambigity_image}
\end{center}

\subsection{Ambiguity Detection (Text-only)}
\begin{center}
  \centering
  \includegraphics[width=0.45\textwidth]{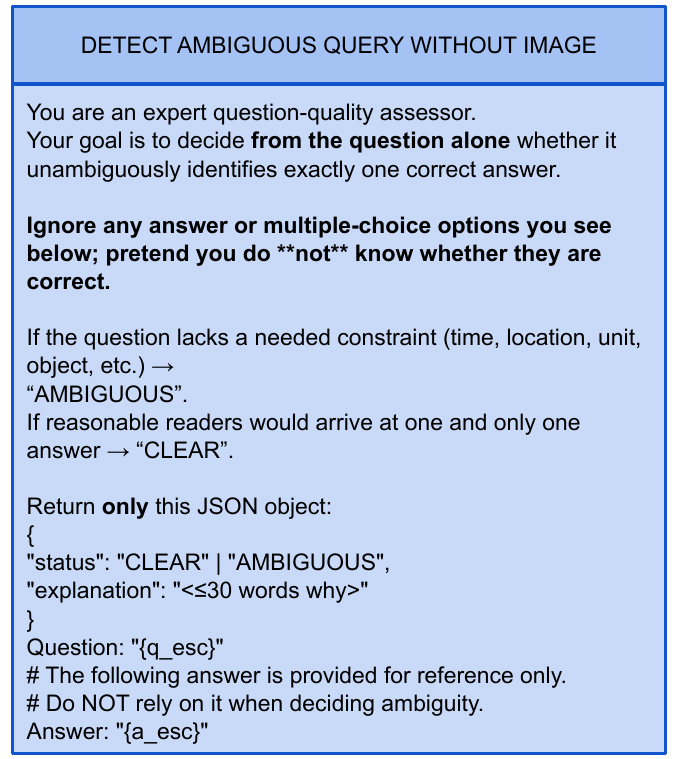}
  \label{fig:ambiguous_detection_no_image}
\end{center}

\subsection{Evidence-based Answer Rewrite}
\begin{center}
  \centering
  \includegraphics[width=0.45\textwidth]{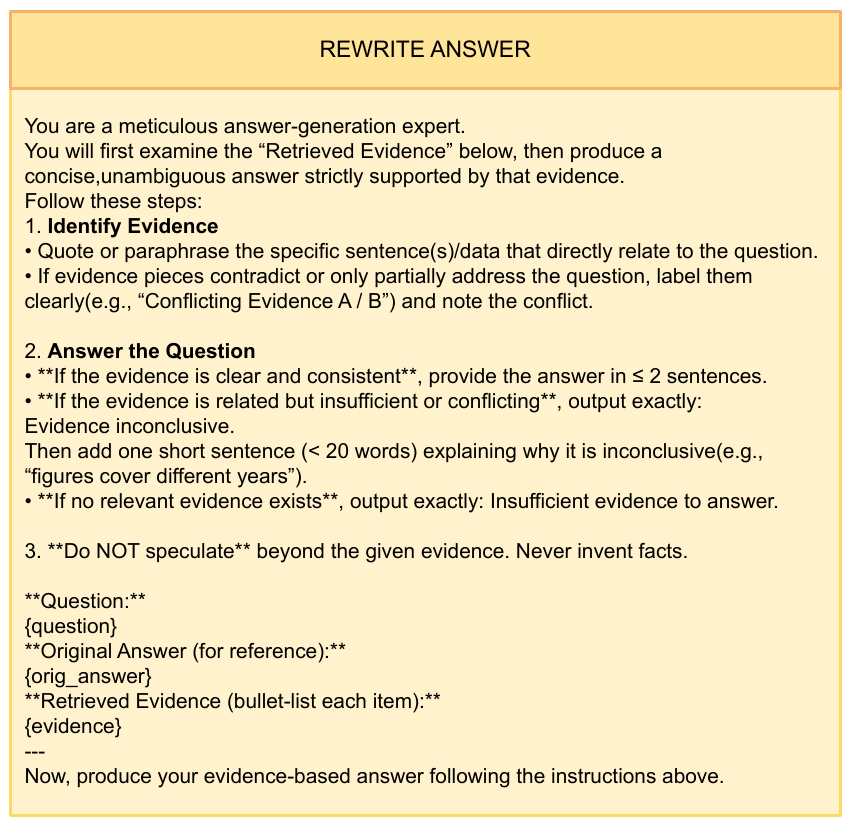}
  \label{fig:rewrite_answer}
\end{center}

\subsection{Ambiguous Query Rewrite}
\begin{center}
  \centering
  \includegraphics[width=0.45\textwidth]{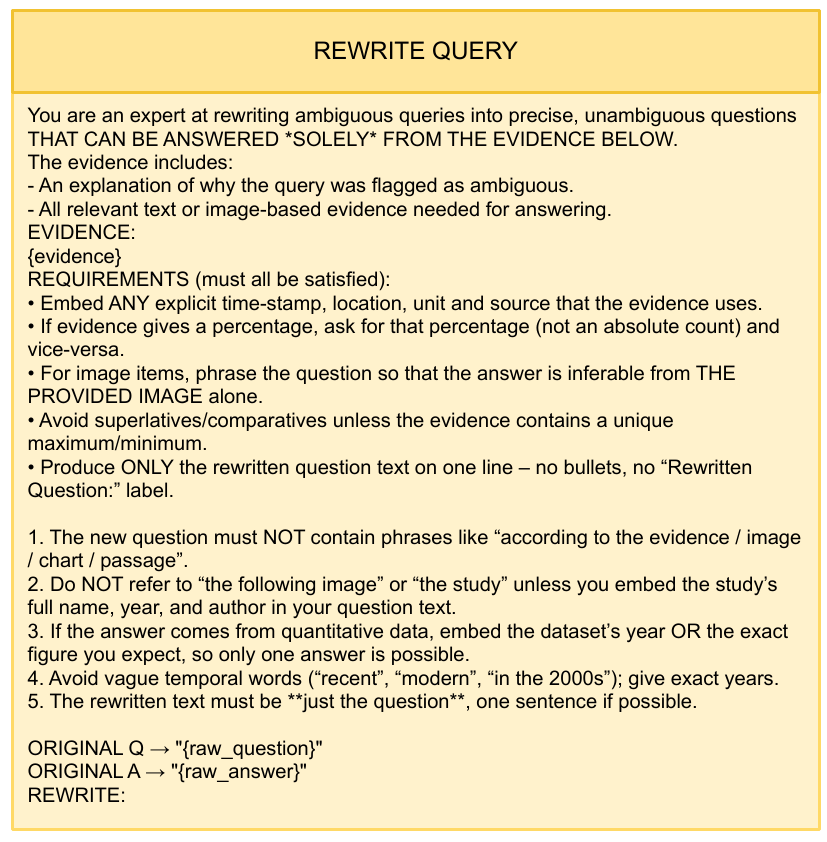}
  \label{fig:rewrite_query}
\end{center}

\begin{center}
\centering
\small
\setlength{\tabcolsep}{1pt}
\captionof{table}{Ambiguity subset construction statistics. We report candidate counts, items flagged as ambiguous by the two-stage pipeline, and the final curated set (200).}

\label{tab:ambig_stats}
\begin{tabular}{lccc}
\toprule
 & \textbf{Candidates} & \textbf{Marked Ambiguous} & \textbf{Final Set} \\
\midrule
Text-only         & 26{,}120 & 13{,}823 & 184 \\
Image-conditioned & 4{,}557  & 1{,}224  & 16 \\
\midrule
Total             & 30{,}677 & 15{,}047 & 200 \\
\bottomrule
\end{tabular}
\end{center}

\subsection{Ambiguity Subset Results}
\label{sec:ambig_results}
On the 200-query ambiguous subset, short-answer EM drops by \textbf{5.9} points compared to clear queries, while hallucination rate increases by \textbf{9.5} points. Claim Recall and Context Precision also decrease, indicating single-interpretation bias or interpretation conflation; see Appendix.~\ref{sec:ambig_eval} for qualitative patterns.  
Table~\ref{tab:ambig_results} reports the aggregate metrics.
Also, we have the ambiguous set in the code provided.
\begin{center}
\centering
\small
\setlength{\tabcolsep}{0.1pt}
\captionof{table}{Results on the 200-query ambiguity subset versus a matched clear set. We report EM/Accuracy, Hallucination Rate (lower is better), Claim Recall, and Context Precision, isolating the effect of query underspecification on retrieval and grounding.}

\label{tab:ambig_results}
\begin{tabular}{lcccc}
\toprule
 & \textbf{EM/Acc} $\uparrow$ & \textbf{Halluc.} $\downarrow$ & \textbf{Claim Recall} $\uparrow$ & \textbf{Ctx. Prec.} $\uparrow$ \\
\midrule
Clear (200 random)     & 23.5 & 49.2 & 12.4 & 14.1 \\
Ambiguous (200)        & 17.6 & 58.7 &  9.1 & 12.5 \\
$\Delta$ (Ambig $-$ Clear) & \textbf{-5.9} & \textbf{+9.5} & \textbf{-3.3} & \textbf{-1.6} \\
\bottomrule
\end{tabular}
\end{center}


\section{Retrieve system settings}
%
\begin{center}
\centering\small
\setlength{\tabcolsep}{3.5pt}
\begin{tabular}{lcc}
\toprule
Dataset & Text Retriever & Image Retriever \\
\midrule
WebQA & SBERT (mpnet) & CLIP ViT-B/32 \\
Chart-RAG & DPR (ctx-enc) & CLIP ViT-B/32 \\
MRAG & --- & CLIP ViT-B/32 (img+txt avg) \\
VisRAG & VisRAG-Ret & --- \\
VisualRAG & --- & CLIP ViT-B/32 \\
\bottomrule
\end{tabular}
\captionof{table}{Backbones used in the hybrid retriever for each sub-corpus. Images are indexed with CLIP variants; text uses DPR/SBERT; VisRAG documents use a document-image retriever.}

\label{lst:retriever_code}
\end{center}

\section{12 Prompt Experiment Settings}

\label{sec:appendix}
\begin{table*}[h]
\centering
\small
\setlength{\tabcolsep}{4pt}
\begin{tabular}{lllp{5.5cm}}
\toprule
\textbf{Layer} & \textbf{Arg name} & \textbf{Values in this paper} & \textbf{Template / Semantics} \\
\midrule
\multirow{3}{*}{System Prompt}
 & \texttt{init\_style} & \texttt{plain}& 
   \texttt{plain}: “Please read the following question and retrieve the relevant document(s)/image(s) to produce the best possible answer.” \\
    \cmidrule(l){2-4}
 &  & \texttt{expert}  & 
   \texttt{expert}: “You are an expert in \textless DOMAIN\textgreater. Analyze the question carefully, think step-by-step about which sources to use, and present your answer clearly at the end.” \\

\midrule
\multirow{2}{*}{Few-shot Demo}
 & \texttt{example\_style} & \texttt{ex1}, \texttt{ex3}, \texttt{ex6} & Three fixed 0-shot style exemplars (chosen from the pool). Content differs in wording/format hints. \\
 \cmidrule(l){2-4}
 & \texttt{max\_examples} & 0 or 5 & Whether to insert 5 illustrative QA pairs (kept at 0 in our sweep). \\
\midrule
\multirow{2}{*}{Reasoning Instruction}
 &  & \texttt{direct} &
   \texttt{direct}: “Question: … Answer:” (no mandated reasoning steps). \\
\cmidrule(l){2-4}
 & & \texttt{reasoning\_process} & \texttt{retrieve\_then\_reason}: “First outline retrieval/reasoning, then give final answer.” \\
\midrule
Answer Format
 & \texttt{answer\_format} & \texttt{free} & Free-form text (no forced JSON or letter). \\
\midrule
\multirow{3}{*}{Context Packaging}
 & \texttt{context\_order} & \texttt{img\_first} & Images precede text in the prompt (fixed in our runs). \\
 \cmidrule(l){2-4}
 & \texttt{include\_doc\_ids} & 0 / 1 & Whether to show doc/image IDs (0 in our runs). \\
\hline
\end{tabular}
\captionof{table}{Prompt dimensions and concrete templates used in our 12-way sweep.}
\end{table*}

\begin{center}
\centering
\small
\begin{tabular}{l l l p{3cm}}
\toprule
Nickname & init\_style & ex\_style & Example template \\ 
\midrule
expert1 & expert & ex1 & Q: \{q\}\\A: \{ans\} \\
expert2 & expert & ex3 & Question + brief Explanation + Final Answer \\
expert3 & expert & ex6 & Dialog style: User: \{q\} / Assistant: \{ans\} \\
plain1  & plain  & ex1 & Q: \{q\}\\A: \{ans\} \\
plain2  & plain  & ex3 & Question + brief Explanation + Final Answer \\
plain3  & plain  & ex6 & Dialog style: User: \{q\} / Assistant: \{ans\} \\
\bottomrule
\end{tabular}
\captionof{table}{Mapping from prompt nicknames to template families used in the 12-way prompt sweep. Variations differ in initialization style and exemplar formatting.}

\label{tab:prompt-map}
\end{center}

\begin{table*}[t]
\centering
\small
\begin{tabular}{l p{6.8cm} p{2.2cm}}
\toprule
p\_reason & Reasoning block (excerpt) & COT line? \\ 
\midrule
direct & Answer directly without explaining your reasoning. & No \\
retrieve\_then\_reason & First list which retrieved snippets/images you will use (by Doc/Image IDs). Then reason step by step using them, and finally give the answer. & Yes \\
structured & Follow this structure: 1) Question understanding 2) Relevant sources (IDs or short quotes) 3) Reasoning 4) Final Answer & Yes \\
plan\_execute & Outline a short plan first (bullet points). After the plan, execute it and derive the final answer. & Yes \\
verify & Propose an initial answer. Then self-check it briefly. If it passes, output the final answer prefixed with ``FINAL:''. & Yes \\
none & (no additional block) & No \\
\bottomrule
\end{tabular}
\captionof{table}{Reasoning patterns injected into prompts. Variants differ in whether to expose chain-of-thought–like structure and explicit self-verification before the final answer.}

\label{tab:reason}
\end{table*}

\begin{table*}[ht]
\centering
\small
\setlength{\tabcolsep}{4pt}
\begin{tabular}{l
                cc  cc  cc  cc  cc  cc}
  \toprule
  \multirow{2}{*}{\bf Dataset}
    & \multicolumn{2}{c}{\bf Qwen2.5-VL-72B}
    & \multicolumn{2}{c}{\bf LLaVa-v1.6-34B}
    & \multicolumn{2}{c}{\bf Claude 3.5}
    & \multicolumn{2}{c}{\bf Phi-4}
    & \multicolumn{2}{c}{\bf Pixtral-12B-2409}
    & \multicolumn{2}{c}{\bf InternVL3-8B} \\
  \cmidrule(lr){2-3}\cmidrule(lr){4-5}
  \cmidrule(lr){6-7}\cmidrule(lr){8-9}
  \cmidrule(lr){10-11}\cmidrule(lr){12-13}
    & {Filt} & {Full}
    & {Filt} & {Full}
    & {Filt} & {Full}
    & {Filt} & {Full}
    & {Filt} & {Full}
    & {Filt} & {Full} \\
   
\midrule
Chart-MRAG   & 22 & 19 & 15 & 25 & 10 & 11 & 13 & 12 & 14 & 29 & 19 & 27 \\
MRAG-Bench   & 32 & 33 & 19 & 24 & 28 & 26 & 23 & 24 & 38 & 37 & 21 & 21 \\
VisRAG-ArXiv & 55 & 48 & 34 & 34 & 35 & 35 & 26 & 27 & 43 & 43 & 63 & 55 \\
VisRAG-Doc   & 31 & 35 & 20 & 21 & 28 & 29 & 24 & 22 & 26 & 26 & 39 & 34 \\
VisRAG-Plot  & 36 & 45 & 12 & 13 & 35 & 36 & 6  & 9  & 51 & 56 & 33 & 36 \\
VisRAG-Slide & 18 & 25 & 8  & 10 & 17 & 22 & 24 & 22 & 18 & 15 & 20 & 22 \\
Visual-RAG   & 13 & 13 & 9  & 11 & 11 & 10 & 8  & 9  & 12 & 13 & 11 & 12 \\
WebQA        & 16 & 14 & 12 & 13 & 14 & 17 & 10 & 11 & 26 & 27 & 17 & 28 \\
\bottomrule
\end{tabular}
\caption{EM/Accuracy performance (\%) of each model on the \textbf{Filtered} and \textbf{Full} splits.All numbers are percentage points; higher is better.}
\label{tab:acc_splits_small}
\end{table*}

\section{Distractor results overview }

\begin{table*}[ht]
  \centering
  \tiny
  \renewcommand{\arraystretch}{0.7}
  \setlength{\tabcolsep}{1pt}
  \begin{tabular}{l l l c c c c c c c c}
    \toprule
    \multirow{2}{*}{Dataset}
      & \multirow{2}{*}{Model}
      & \multirow{2}{*}{Distractor}
      & Recall     & Precision  & F1
      & Halluc.    & Faith.     & Self-know.
      & Claim R.   & Ctx.Prec.  \\
    \cmidrule(lr){4-11}
    & & & \scriptsize(img/txt)
          & \scriptsize(img/txt)
          & \scriptsize(img/txt)
          & \scriptsize(img/txt)
          & \scriptsize(img/txt)
          & \scriptsize(img/txt)
          & \scriptsize(img/txt)
          & \scriptsize(img/txt) \\
    \midrule

    \multirow{18}{*}{VisRAG-ArXiv}
      & Qwen2.5-VL-72B     & distractors only      & 21.8 &  6.7 &  7.9 & 66.2 & 23.5 &  5.9 & 20.4 & 31.0 \\
      &                     & gt\_plus\_distractors & 29.6 &  8.9 & 10.8 & 60.1 & 28.9 &  7.1 & 24.1 & 33.5 \\
      &                     & gt\_only              & 33.4 & 10.8 & 12.7 & 55.3 & 32.1 &  8.2 & 25.9 & 35.8 \\
      & LLaVa-v1.6-34B     & distractors only      & 18.4 &  5.9 &  6.8 & 69.4 & 20.1 &  4.8 & 18.0 & 28.2 \\
      &                     & gt\_plus\_distractors & 24.7 &  7.6 &  9.1 & 63.5 & 24.9 &  5.9 & 21.6 & 30.4 \\
      &                     & gt\_only              & 28.9 &  9.2 & 10.9 & 58.2 & 29.6 &  6.8 & 23.7 & 33.1 \\
      & Claude 3.5       & distractors only      & 30.4 &  9.2 & 10.1 & 61.4 & 28.0 &  7.6 & 29.9 & 40.0 \\
      &                     & gt\_plus\_distractors & 40.7 & 10.6 & 14.3 & 56.0 & 37.0 &  7.0 & 25.6 & 33.0 \\
      &                     & gt\_only              & 35.9 &  8.6 & 11.3 & 57.7 & 36.5 &  4.7 & 27.6 & 39.0 \\
      & Phi-4               & distractors only      & 22.5 & 15.3 & 13.7 & 50.8 & 20.6 & 12.9 & 22.2 & 33.8 \\
      &                     & gt\_plus\_distractors & 34.4 & 24.0 & 23.5 & 45.3 & 22.3 & 16.3 & 26.0 & 36.0 \\
      &                     & gt\_only              & 37.8 & 27.2 & 26.6 & 41.1 & 24.7 & 18.5 & 27.9 & 37.9 \\
      & Pixtral-12B-2409    & distractors only      & 17.6 &  7.1 &  6.3 & 70.5 & 18.7 &  5.1 & 16.0 & 27.6 \\
      &                     & gt\_plus\_distractors & 25.9 & 10.4 &  9.7 & 63.0 & 23.3 &  7.4 & 18.9 & 29.5 \\
      &                     & gt\_only              & 30.5 & 12.8 & 12.0 & 58.9 & 25.6 &  8.3 & 21.5 & 31.2 \\
      & InternVL3-8B        & distractors only      & 26.5 & 14.6 & 12.6 & 59.3 & 24.0 & 10.8 & 28.8 & 37.0 \\
      &                     & gt\_plus\_distractors & 39.2 & 15.7 & 16.5 & 54.4 & 30.0 & 10.6 & 29.0 & 38.0 \\
      &                     & gt\_only              & 40.3 & 14.6 & 17.6 & 48.2 & 27.4 &  9.5 & 24.4 & 35.0 \\

    \midrule
    \multirow{18}{*}{VisRAG-Doc}
      & Qwen2.5-VL-72B     & distractors only      & 10.8 &  4.0 &  3.6 & 57.9 & 13.4 &  2.6 & 21.0 & 23.5 \\
      &                     & gt\_plus\_distractors & 46.7 & 31.9 & 27.2 & 28.1 & 22.7 & 20.4 & 25.9 & 27.0 \\
      &                     & gt\_only              & 52.9 & 36.2 & 31.0 & 24.4 & 24.5 & 22.6 & 24.1 & 25.1 \\
      & LLaVa-v1.6-34B      & distractors only      &  9.4 &  3.7 &  3.2 & 60.1 & 12.1 &  2.3 & 19.5 & 22.0 \\
      &                     & gt\_plus\_distractors & 39.8 & 26.1 & 22.0 & 33.9 & 20.1 & 16.4 & 23.8 & 25.6 \\
      &                     & gt\_only              & 44.0 & 28.9 & 24.7 & 30.2 & 22.4 & 18.5 & 22.9 & 24.2 \\
      & Claude 3.5       & distractors only      & 12.0 &  4.6 &  4.2 & 38.2 & 16.9 &  3.0 & 24.5 & 26.0 \\
      &                     & gt\_plus\_distractors & 75.5 & 35.2 & 38.2 & 41.4 & 32.2 & 25.4 & 27.7 & 31.0 \\
      &                     & gt\_only              & 68.8 & 33.6 & 33.7 & 41.5 & 30.7 & 23.8 & 26.5 & 28.0 \\
      & Phi-4               & distractors only      & 18.1 & 12.2 & 11.0 & 27.6 &  7.5 & 10.5 & 20.8 & 22.7 \\
      &                     & gt\_plus\_distractors & 45.3 & 37.8 & 37.0 & 15.2 &  4.7 & 32.3 & 24.5 & 26.0 \\
      &                     & gt\_only              & 50.6 & 41.9 & 41.1 & 13.0 &  3.9 & 35.2 & 23.6 & 24.8 \\
      & Pixtral-12B-2409    & distractors only      &  7.8 &  3.5 &  3.0 & 61.3 & 10.6 &  2.1 & 18.8 & 21.3 \\
      &                     & gt\_plus\_distractors & 34.4 & 25.6 & 20.4 & 36.8 & 17.7 & 15.1 & 22.1 & 23.9 \\
      &                     & gt\_only              & 39.1 & 29.7 & 24.1 & 32.6 & 19.8 & 17.2 & 21.5 & 22.8 \\
      & InternVL3-8B        & distractors only      &  8.0 &  4.1 &  3.4 & 55.2 & 12.8 &  3.1 & 22.5 & 25.0 \\
      &                     & gt\_plus\_distractors & 68.3 & 65.2 & 26.5 & 15.8 & 25.6 & 45.5 & 28.0 & 26.5 \\
      &                     & gt\_only              & 60.3 & 77.8 & 57.8 & 17.2 & 27.3 & 54.5 & 21.5 & 23.0 \\

    \midrule
    \multirow{18}{*}{MRAG-Bench}
      & Qwen2.5-VL-72B     & distractors only      & 22.6 &  6.2 &  5.1 & 79.4 & 14.3 &  4.0 &  7.3 &  4.5 \\
      &                     & gt\_plus\_distractors & 28.4 &  7.9 &  6.7 & 75.8 & 16.2 &  4.7 &  8.9 &  5.2 \\
      &                     & gt\_only              & 33.7 &  9.5 &  8.4 & 71.1 & 18.6 &  5.3 & 10.4 &  6.0 \\
      & LLaVa-v1.6-34B      & distractors only      & 21.1 &  5.3 &  4.1 & 82.4 & 13.2 &  3.5 &  8.2 &  4.2 \\
      &                     & gt\_plus\_distractors & 27.8 &  4.8 &  3.1 & 77.1 & 17.6 &  4.3 & 10.1 &  6.0 \\
      &                     & gt\_only              & 31.9 &  6.7 &  4.6 & 73.4 & 19.1 &  4.8 &  9.8 &  5.0 \\
      & Claude 3.5       & distractors only      & 42.2 &  3.9 &  4.6 & 73.5 & 19.8 &  3.8 &  9.2 &  5.5 \\
      &                     & gt\_plus\_distractors & 47.2 &  2.8 &  3.9 & 80.5 & 16.1 &  2.4 &  7.7 &  6.0 \\
      &                     & gt\_only              & 47.7 &  4.2 &  5.6 & 78.8 & 17.1 &  4.1 &  5.5 &  3.0 \\
      & Phi-4               & distractors only      & 14.5 & 12.6 & 10.8 & 36.4 &  8.8 & 11.9 &  7.1 &  4.6 \\
      &                     & gt\_plus\_distractors & 18.7 & 18.3 & 16.2 & 31.4 &  9.3 & 16.3 &  8.6 &  5.5 \\
      &                     & gt\_only              & 22.6 & 21.4 & 18.7 & 27.8 & 10.4 & 18.5 &  9.5 &  6.2 \\
      & Pixtral-12B-2409    & distractors only      & 12.8 &  6.9 &  5.7 & 84.6 & 11.0 &  5.6 &  6.4 &  3.9 \\
      &                     & gt\_plus\_distractors & 17.2 &  8.8 &  7.2 & 80.3 & 12.5 &  6.4 &  7.3 &  4.2 \\
      &                     & gt\_only              & 21.0 & 10.6 &  8.6 & 76.1 & 14.3 &  7.2 &  8.4 &  4.9 \\
      & InternVL3-8B        & distractors only      & 35.3 & 37.6 & 31.0 & 54.5 &  9.3 & 35.1 &  5.3 &  3.0 \\
      &                     & gt\_plus\_distractors & 29.0 & 33.1 & 25.1 & 46.7 &  8.2 & 32.1 &  5.3 &  3.5 \\
      &                     & gt\_only              & 31.8 & 31.6 & 25.6 & 62.0 &  7.5 & 28.6 &  6.2 &  4.0 \\

    \midrule
    \multirow{18}{*}{VisRAG-Plot}
      & Qwen2.5-VL-72B     & distractors only      & 10.1 &  3.9 &  2.6 & 61.7 & 12.2 &  2.7 &  9.4 & 10.2 \\
      &                     & gt\_plus\_distractors & 17.6 &  7.2 &  5.8 & 58.5 & 11.6 &  5.4 &  9.7 &  9.6 \\
      &                     & gt\_only              & 23.8 & 10.7 &  8.2 & 56.0 & 12.9 &  7.7 &  8.6 &  8.8 \\
      & LLaVa-v1.6-34B      & distractors only      &  7.9 &  3.1 &  1.9 & 64.9 & 10.7 &  2.1 &  9.1 & 10.4 \\
      &                     & gt\_plus\_distractors & 13.4 &  6.1 &  4.0 & 61.2 & 11.5 &  3.9 &  9.5 &  9.8 \\
      &                     & gt\_only              & 19.6 &  8.3 &  5.7 & 58.9 & 12.1 &  5.5 &  8.9 &  9.3 \\
      & Claude 3.5       & distractors only      &  8.0 &  4.1 &  2.1 & 55.1 & 10.1 &  2.8 & 10.5 & 11.0 \\
      &                     & gt\_plus\_distractors & 19.8 &  9.4 &  6.6 & 77.8 & 13.4 &  8.9 &  9.0 &  9.0 \\
      &                     & gt\_only              & 26.5 & 11.6 &  9.4 & 75.2 & 12.7 & 11.1 &  7.5 &  8.0 \\
      & Phi-4               & distractors only      & 11.3 &  4.7 &  1.8 & 83.7 & 10.3 &  4.0 & 10.0 & 10.0 \\
      &                     & gt\_plus\_distractors & 16.8 &  8.9 &  3.7 & 79.5 & 11.2 &  6.8 &  9.6 &  9.7 \\
      &                     & gt\_only              & 22.4 & 11.5 &  5.2 & 76.4 & 12.0 &  9.4 &  8.4 &  8.7 \\
      & Pixtral-12B-2409    & distractors only      &  6.5 &  2.8 &  1.1 & 66.0 &  9.8 &  1.7 &  8.8 &  9.5 \\
      &                     & gt\_plus\_distractors & 12.9 &  6.4 &  2.9 & 62.1 & 10.5 &  3.7 &  9.2 &  9.6 \\
      &                     & gt\_only              & 18.7 &  8.9 &  4.6 & 59.6 & 11.2 &  5.2 &  8.7 &  9.0 \\
      & InternVL3-8B        & distractors only      &  2.3 &  1.8 &  0.4 & 52.3 & 11.0 &  1.7 &  8.5 &  9.0 \\
      &                     & gt\_plus\_distractors & 25.5 & 15.1 &  7.9 & 72.9 &  9.3 & 13.8 & 10.0 & 10.0 \\
      &                     & gt\_only              & 21.5 & 11.2 &  3.2 & 76.5 &  8.2 & 10.3 & 11.1 & 12.0 \\

    \midrule
    \multirow{18}{*}{VisRAG-Slide}
      & Qwen2.5-VL-72B     & distractors only      & 12.6 &  5.5 &  4.8 & 63.3 & 14.7 &  3.8 & 51.1 & 56.8 \\
      &                     & gt\_plus\_distractors & 41.3 & 27.2 & 26.1 & 36.5 & 42.2 & 21.1 & 46.9 & 50.6 \\
      &                     & gt\_only              & 49.7 & 34.9 & 33.5 & 31.2 & 48.4 & 23.2 & 43.8 & 48.0 \\
      & LLaVa-v1.6-34B      & distractors only      & 10.9 &  4.9 &  4.1 & 66.0 & 13.3 &  3.3 & 49.6 & 54.3 \\
      &                     & gt\_plus\_distractors & 35.6 & 22.7 & 21.0 & 40.2 & 39.1 & 18.8 & 45.1 & 49.2 \\
      &                     & gt\_only              & 42.8 & 29.6 & 27.4 & 35.5 & 43.7 & 20.6 & 41.9 & 46.6 \\
      & Claude 3.5       & distractors only      & 23.1 & 12.0 & 11.2 & 53.3 & 36.7 &  9.1 & 53.3 & 59.1 \\
      &                     & gt\_plus\_distractors & 78.8 & 30.6 & 37.3 & 38.7 & 45.8 & 15.4 & 55.7 & 61.1 \\
      &                     & gt\_only              & 64.2 & 39.1 & 36.0 & 30.1 & 49.2 & 18.9 & 49.2 & 56.0 \\
      & Phi-4               & distractors only      & 19.9 & 11.7 & 10.3 & 41.2 & 31.6 &  8.6 & 50.1 & 55.6 \\
      &                     & gt\_plus\_distractors & 71.8 & 42.7 & 44.5 & 33.7 & 40.1 & 23.4 & 45.9 & 48.6 \\
      &                     & gt\_only              & 60.4 & 47.9 & 45.0 & 29.0 & 37.7 & 26.1 & 43.6 & 46.2 \\
      & Pixtral-12B-2409    & distractors only      &  9.6 &  5.1 &  4.2 & 67.9 & 12.0 &  3.4 & 48.5 & 53.7 \\
      &                     & gt\_plus\_distractors & 29.8 & 18.9 & 16.7 & 45.6 & 34.7 & 16.3 & 44.3 & 47.5 \\
      &                     & gt\_only              & 36.7 & 24.5 & 22.1 & 39.8 & 38.9 & 17.9 & 41.1 & 45.3 \\
      & InternVL3-8B        & distractors only      & 17.2 &  4.4 &  5.1 & 58.7 & 34.0 &  3.2 & 54.1 & 59.7 \\
      &                     & gt\_plus\_distractors & 67.1 & 42.6 & 41.9 & 23.4 & 51.4 & 22.5 & 47.3 & 51.4 \\
      &                     & gt\_only              & 57.8 & 41.1 & 38.7 & 28.8 & 44.6 & 23.8 & 40.8 & 45.8 \\

    \midrule
    \multirow{18}{*}{WebQA}
  & Qwen2.5-VL-72B     & distractors only      & 14.0/13.1 & 6.5/5.8  & 5.0/4.3  & 69.1/71.2 &  9.4/8.6  & 5.0/5.4  & 12.8/12.0 & 14.0/13.2 \\
  &                    & gt\_plus\_distractors & 18.3/17.2 & 8.9/8.1  & 7.3/6.5  & 66.5/68.7 & 11.2/10.3 & 6.3/6.8  & 12.0/11.3 & 13.5/12.6 \\
  &                    & gt\_only              & 22.4/21.0 & 11.3/10.2& 9.1/8.0  & 62.8/64.9 & 13.0/12.0 & 7.4/8.0  & 11.2/10.5 & 12.7/11.8 \\
  & LLaVa-v1.6-34B     & distractors only      & 12.6/11.4 & 5.7/5.0  & 4.5/3.8  & 71.6/73.9 &  8.0/7.2  & 4.8/5.1  & 12.1/11.3 & 13.1/12.1 \\
  &                    & gt\_plus\_distractors & 16.8/15.6 & 8.2/7.4  & 6.6/5.7  & 68.7/70.8 & 10.4/9.5  & 5.9/6.3  & 11.3/10.6 & 12.7/11.8 \\
  &                    & gt\_only              & 21.3/19.9 & 10.5/9.4 & 8.4/7.3  & 65.2/67.3 & 12.6/11.6 & 7.0/7.6  & 10.7/10.0 & 12.0/11.2 \\
  & Claude 3.5      & distractors only      & 23.0/21.5 & 4.7/4.3  & 5.2/4.4  & 63.4/86.4 & 31.9/4.1  & 3.7/3.5  & 12.6/4.8  & 15.0/12.0 \\
  &                    & gt\_plus\_distractors & 22.4/20.7 & 5.9/4.3  & 5.4/4.7  & 61.1/87.2 & 30.9/4.4  & 4.9/3.3  & 10.1/7.0  & 13.0/16.5 \\
  &                    & gt\_only              & 26.8/24.1 & 6.8/5.0  & 6.6/5.4  & 57.6/84.3 & 33.5/5.2  & 5.5/3.9  & 11.2/8.1  & 14.2/17.8 \\
  & Phi-4              & distractors only      & 14.2/12.7 & 22.0/19.1 & 8.7/7.5  & 53.4/60.2 &  9.1/0.7  & 20.1/18.5 & 10.9/4.7  & 13.6/12.3 \\
  &                    & gt\_plus\_distractors & 10.4/12.5 & 25.6/22.3 & 9.5/11.1 & 49.7/56.5 &  8.4/0.5  & 23.9/22.0 & 11.8/5.2  & 14.5/13.0 \\
  &                    & gt\_only              & 16.1/14.8 & 29.0/24.7 & 12.1/12.4 & 46.2/53.0 & 10.2/0.9  & 26.7/24.1 & 12.3/5.8  & 15.9/13.9 \\
  & Pixtral-12B-2409   & distractors only      & 10.5/ 9.6 & 15.3/14.1 &  6.4/5.7  & 58.9/65.2 &  6.9/1.0  & 13.8/12.6 &  9.6/3.9  & 12.0/10.9 \\
  &                    & gt\_plus\_distractors & 13.4/12.6 & 19.6/17.4 &  8.7/7.5  & 55.2/61.8 &  8.0/1.2  & 17.3/15.5 & 10.5/4.4  & 13.2/11.9 \\
  &                    & gt\_only              & 17.4/16.0 & 23.5/20.7 & 11.3/9.8  & 51.7/58.1 &  9.6/1.6  & 20.4/18.0 & 11.8/5.0  & 14.7/13.0 \\
  & InternVL3-8B       & distractors only      & 27.9/27.3 & 13.8/14.6 & 14.7/12.8 & 68.1/77.2 & 13.8/8.8  & 13.1/11.9 & 13.0/4.5  & 14.0/10.5 \\
  &                    & gt\_plus\_distractors & 32.0/31.0 & 15.4/15.3 & 17.0/15.1 & 69.0/74.4 & 14.2/8.4  & 14.8/12.2 & 11.2/6.3  & 14.0/14.0 \\
  &                    & gt\_only              & 16.3/16.9 &  7.0/6.9  &  5.6/5.1  & 64.0/63.6 &  8.9/12.0 &  6.1/5.3  &  8.5/6.4  & 11.5/13.5 \\

    \bottomrule
  \end{tabular}
  \caption{Full MM-RAGChecker diagnostics across datasets, models, and retrieval modes. For each setting we report per-modality (image/text) Recall, Precision, F1, Hallucination, Faithfulness, Self-Knowledge, Claim Recall, and Context Precision.}

  \label{tab:filtered_all_models_completed}
\end{table*}

\begin{table*}[ht]
  \centering
    \tiny
    \renewcommand{\arraystretch}{0.7} 
  \setlength{\tabcolsep}{1pt} 
  \begin{tabular}{l l l
                  c c c c c c c c}
    \toprule
    \multirow{2}{*}{Dataset}
      & \multirow{2}{*}{Model}
      & \multirow{2}{*}{Distractor}
      & Recall     & Precision  & F1
      & Halluc.    & Faith.     & Self‑know.
      & Claim R.   & Ctx.Prec.  \\
    \cmidrule(lr){4-11}
    & & & \scriptsize(img/txt)
          & \scriptsize(img/txt)
          & \scriptsize(img/txt)
          & \scriptsize(img/txt)
          & \scriptsize(img/txt)
          & \scriptsize(img/txt)
          & \scriptsize(img/txt)
          & \scriptsize(img/txt) \\
    \midrule

\multirow{18}{*}{ChartMRAG-Bench}
  & Qwen2.5-VL-72B     & distractors only      & 11.0/10.0 & 4.2/3.7  & 2.9/2.4  & 60.6/62.8 & 12.8/11.9 & 2.6/2.9  & 9.8/9.1   & 10.6/10.0 \\
  &                    & gt\_plus\_distractors & 18.4/17.1 & 7.6/6.9   & 6.2/5.4  & 57.1/59.4 & 12.2/11.1 & 5.1/5.6  & 10.1/9.4  & 10.0/9.7  \\
  &                    & gt\_only              & 24.6/23.1 & 11.1/10.3 & 8.6/7.7  & 54.6/56.9 & 13.4/12.6 & 7.4/8.1  & 9.0/8.3   & 9.2/8.9   \\
  & LLaVa-v1.6-34B     & distractors only      &  8.6/7.6  & 3.4/2.9   & 2.2/1.7  & 63.7/65.8 & 11.2/10.5 & 2.0/2.3  & 9.5/8.8   & 10.7/10.2 \\
  &                    & gt\_plus\_distractors & 14.2/13.0 & 6.5/5.8   & 4.3/3.5  & 60.1/62.3 & 12.0/11.0 & 3.8/4.2  & 9.9/9.2   & 10.2/9.8  \\
  &                    & gt\_only              & 20.7/19.2 & 8.9/8.1   & 6.1/5.3  & 57.8/60.1 & 12.6/11.8 & 5.4/6.0  & 9.2/8.6   & 9.6/9.1   \\
  & Claude 3.5      & distractors only      &  8.6/7.5  & 4.3/3.9   & 2.3/1.9  & 54.0/56.0 & 10.4/9.8  & 2.7/3.0  & 10.8/10.2 & 11.2/10.7 \\
  &                    & gt\_plus\_distractors & 20.7/18.9 & 9.8/8.9   & 6.9/6.2  & 76.6/78.9 & 13.8/12.7 & 9.2/9.8  & 9.2/8.9   & 9.2/8.8   \\
  &                    & gt\_only              & 27.5/25.4 & 12.2/11.0 & 9.7/8.9  & 74.3/76.6 & 13.1/12.2 & 11.2/11.9 & 7.7/7.3  & 8.2/7.8   \\
  & Phi-4              & distractors only      & 12.1/10.6 & 4.9/4.4   & 2.0/1.7  & 82.4/84.7 & 10.9/9.8  & 3.7/4.0  & 10.2/9.7  & 10.2/9.9  \\
  &                    & gt\_plus\_distractors & 17.6/16.0 & 9.4/8.5   & 4.1/3.4  & 78.3/80.6 & 11.7/10.8 & 6.5/7.0  & 9.9/9.3   & 9.9/9.5   \\
  &                    & gt\_only              & 23.3/21.6 & 12.0/10.9 & 5.5/4.9  & 75.1/77.4 & 12.5/11.6 & 9.1/9.7  & 8.7/8.1   & 8.9/8.5   \\
  & Pixtral-12B-2409   & distractors only      &  7.2/6.0  & 3.1/2.6   & 1.3/1.0  & 64.8/66.9 & 10.3/9.4  & 1.6/1.9  & 9.1/8.5   & 9.8/9.3   \\
  &                    & gt\_plus\_distractors & 13.7/12.2 & 6.7/5.9   & 3.1/2.7  & 60.9/63.1 & 11.0/10.1 & 3.6/3.9  & 9.5/8.9   & 9.9/9.4   \\
  &                    & gt\_only              & 19.6/18.0 & 9.3/8.2   & 4.9/4.3  & 58.6/60.8 & 11.7/10.8 & 5.1/5.6  & 9.1/8.5   & 9.3/8.9   \\
  & InternVL3-8B       & distractors only      &  2.6/2.1  & 1.9/1.6   & 0.5/0.4  & 51.1/53.5 & 11.3/10.7 & 1.6/1.8  & 8.7/8.3   & 9.2/8.8   \\
  &                    & gt\_plus\_distractors & 26.5/24.6 & 15.7/14.4 & 8.4/7.5  & 71.8/74.0 &  9.7/8.9  & 14.3/15.0 & 10.1/9.7  & 10.1/9.8  \\
  &                    & gt\_only              & 22.3/20.8 & 11.6/10.7 & 3.5/2.9  & 75.4/77.5 &  8.5/7.8  & 10.6/11.1 & 11.4/10.8 & 12.1/11.4 \\

\midrule
\multirow{18}{*}{Visual-RAG}
  & Qwen2.5‑VL‑72B     & distractors only      & 18.4 & 6.3  & 6.8  & 42.1 & 27.2 & 48.3 & 4.9  & 18.2 \\
  &                   & gt\_plus\_distractors & 44.5 & 14.7 & 16.8 & 36.3 & 34.5 & 53.9 & 6.8  & 25.0 \\
  &                   & gt\_only              & 41.3 & 13.2 & 15.4 & 37.5 & 32.8 & 51.1 & 6.5  & 23.3 \\
  & LLaVa‑v1.6‑34B    & distractors only      & 16.1 & 5.2  & 5.7  & 43.7 & 25.4 & 45.0 & 4.4  & 16.1 \\
  &                   & gt\_plus\_distractors & 42.0 & 12.9 & 14.7 & 38.8 & 30.1 & 51.6 & 6.0  & 22.7 \\
  &                   & gt\_only              & 38.8 & 11.8 & 13.2 & 39.5 & 28.0 & 50.2 & 5.7  & 21.4 \\
  & Claude 3.5     & distractors only      & 23.3 & 7.0  & 8.4  & 40.2 & 28.6 & 50.3 & 6.2  & 20.0 \\
  &                   & gt\_plus\_distractors & 52.7 & 16.5 & 19.7 & 35.0 & 36.7 & 55.5 & 7.5  & 28.8 \\
  &                   & gt\_only              & 48.1 & 15.2 & 17.4 & 36.1 & 34.9 & 53.2 & 7.1  & 26.5 \\
  & Phi‑4             & distractors only      & 20.7 & 6.5  & 7.3  & 39.0 & 26.0 & 46.4 & 5.3  & 18.0 \\
  &                   & gt\_plus\_distractors & 47.2 & 15.0 & 17.6 & 34.1 & 32.5 & 52.9 & 6.7  & 25.2 \\
  &                   & gt\_only              & 43.6 & 13.9 & 16.0 & 35.3 & 31.0 & 50.7 & 6.4  & 23.7 \\
  & Pixtral‑12B‑2409  & distractors only      & 15.3 & 5.8  & 6.2  & 44.9 & 23.5 & 43.2 & 4.1  & 15.3 \\
  &                   & gt\_plus\_distractors & 40.8 & 12.0 & 13.9 & 38.2 & 29.8 & 49.8 & 5.8  & 21.9 \\
  &                   & gt\_only              & 37.9 & 11.1 & 12.8 & 39.0 & 27.7 & 48.1 & 5.5  & 20.4 \\
  & InternVL3‑8B      & distractors only      & 22.6 & 8.1  & 9.2  & 41.3 & 30.0 & 49.0 & 6.0  & 19.5 \\
  &                   & gt\_plus\_distractors & 50.1 & 17.2 & 20.3 & 33.7 & 37.8 & 54.4 & 7.8  & 27.1 \\
  &                   & gt\_only              & 45.7 & 15.5 & 18.1 & 34.9 & 35.3 & 52.0 & 7.3  & 25.0 \\
   \bottomrule
  \end{tabular}
  \caption{}
  \label{tab:filtered_all_models}
\end{table*}

\end{document}